\pdfoutput=1

\documentclass[11pt]{article}

\usepackage[]{ACL2023}

\usepackage{times}
\usepackage{latexsym}

\usepackage[T1]{fontenc}

\usepackage[utf8]{inputenc}

\usepackage{microtype}

\usepackage{url}
\usepackage{multirow}
\usepackage{booktabs}
\usepackage{tabularx}
\usepackage{graphics}
\usepackage{graphicx}
\usepackage{colortbl}
\usepackage{xcolor,soul}
\usepackage{makecell}
\usepackage{amsmath}

\usepackage[label font=bf]{subfig}
\usepackage[bulgarian, english]{babel}
\usepackage{CJKutf8}
\newcommand{\zh}[1]{\begin{CJK}{UTF8}{bsmi}#1\end{CJK}}

\newcommand\authorfootnote[1]{%
  \begingroup
  \renewcommand\thefootnote{}\footnote{#1}%
  \addtocounter{footnote}{-1}%
  \endgroup
}

\usepackage{tcolorbox}
\usepackage{adjustbox}
\definecolor{azure}{rgb}{0.0, 0.5, 1.0}

\title{Translate to Disambiguate: Zero-shot Multilingual Word Sense Disambiguation with Pretrained Language Models}

\author{Haoqiang Kang$^{*}$ \quad Terra Blevins$^{*}$ \quad Luke Zettlemoyer \\
        Paul G. Allen School of Computer Science \& Engineering,\\ University of Washington \\
        {\tt \{haoqik, blvns, lsz\}@cs.washington.edu}}

\begin{document}
\maketitle
\authorfootnote{$^*$These authors contributed equally to this work.}

\begin{abstract}
Pretrained Language Models (PLMs) learn rich cross-lingual knowledge and can be finetuned to perform well on diverse tasks such as translation and multilingual word sense disambiguation (WSD). However, they often struggle at disambiguating word sense in a zero-shot setting. To better understand this contrast, we present a new study investigating how well PLMs capture cross-lingual word sense with Contextual Word-Level Translation (C-WLT), an extension of word-level translation that prompts the model to translate a given word in context. We find that as the model size increases, PLMs encode more cross-lingual word sense knowledge and better use context to improve WLT performance. Building on C-WLT, we introduce a zero-shot approach for WSD, tested on 18 languages from the XL-WSD dataset. Our method outperforms fully supervised baselines on recall for many evaluation languages without additional training or finetuning. This study presents a first step towards understanding how to best leverage the cross-lingual knowledge inside PLMs for robust zero-shot reasoning in any language.
\end{abstract}

\section{Introduction}

Pretrained Language Models (PLMs) have been found to perform many cross-lingual tasks without explicit cross-lingual training signals, including word-level translation (WLT) across languages \cite{gonen2020greek}. These models also demonstrate cross-lingual knowledge when finetuned for the word sense disambiguation (WSD) \cite{raganato-etal-2020-xl, pasini2021xl}. However, little is known about the extent to which word sense knowledge comes from pretraining rather than finetuning: many PLMs struggle to disambiguate word sense when formulated as a binary classification task, the most common word sense setup for prompting language models \cite{shi2022language, scao2022bloom}.

To investigate this, we measure the ability of multilingual autoregressive language models to understand the cross-lingual meaning of words in a given context. Specifically, we extend the WLT task setup to include a specific context in the prompt, which we call Contextual Word-Level Translation (C-WLT).
We show empirically that pretrained language models are able to take advantage of contextual information in the prompt to improve WLT performance, and as the model size increases, both English and multilingual PLM demonstrate improved cross-lingual knowledge resulting in better performance in contextual WLT.

Translations of a word that change based on context are frequently due to differing word senses not shared by an analogous word in the target language \cite{resnik1999distinguishing}. Inspired by this, we apply C-WLT to the task of WSD by translating the ambiguous word $w$ in context with WLT and then assigning $w$ with the senses in the overlap of the translated word's sense set with $w$'s senses (Figure \ref{fig:xl-wsd-method}, left). We test this zero-shot approach for WSD on 18 languages from the XL-WSD dataset \cite{pasini2021xl}, and find that in our best setting, WSD via C-WLT outperforms prior works on recall for many evaluation languages with no additional training or finetuning of the model. We also observe that ensembling diverse target languages with this method narrows down the predicted set of senses, as demonstrated by the improvements in Jaccard similarity with the reference set. Finally, we analyze our design choices and the types of errors made by this approach to better understand the behavior of WSD via C-WLT and how it relates to supervised WSD classification.

The overall findings of this work are as follows:

\begin{itemize}
\itemsep0em

\item PLMs leverage contextual information to encode cross-lingual knowledge and better capture lexical information, such as word translations and meanings.

\item We can leverage this contextual knowledge of lexical translation to effectively perform zero-shot WSD for many languages, including low-resource ones and languages on which the PLM was not pretrained.

\item The efficacy of WSD via C-WLT depends on the interplay between pretraining languages, model size, and target language choice: smaller multilingual PLMs perform better on seen languages but are more sensitive to design choices and do not generalize as well as larger English PLMs.
 
\end{itemize}

\noindent In sum, we evaluate the lexical translation skills of PLMs in context, and we present a first step towards applying that skill to the downstream task of WSD. Given that most WSD training data outside of English are automatically created \cite[e.g.,][]{scarlini2019just, barba2021mulan}, zero-shot approaches such as our proposed WSD via C-WLT approach are crucial for improving WSD in lower-resource languages.

\section{Contextual Word-Level Translation}

A common method of evaluating the cross-lingual capabilities of PLMs is the task of a word-level translation (WLT), where the model is prompted to translate a word $w_s$ from a source language $L_s$ into another target language $L_t$ \cite{gonen2020greek}. However, this setup does not consider variations in the translation of $w_s$ into $L_t$ that occur when the surface form of $w_s$ represents multiple meanings, or senses, in different contexts.

We propose an extension of the word-level translation task, Contextual Word-Level Translation (C-WLT), which requires translating words correctly based on how they are used in a given context (Figure \ref{fig:xl-wsd-method}, right panel). Specifically, we prompt the PLM to translate $w_s$ from $L_s$ into $L_t$ when conditioned on a specific context $c_s$ where $w_s \in c_s$; we then measure whether it produced the correct translation(s) $w_t$ in context of $w_s$. 

For example, if we want to translate ``plant'' into Chinese based on the context sentence ``The plant sprouted a new leaf'', we prompt the PLM with \textit{In the sentence ``The plant sprouted a new leaf'', the word ``plant'' is translated into Chinese as \_\_}. This evaluation allows us to quantify a PLM's ability to align meaning across languages in a context-specific manner.

\subsection{Experimental Setup}

\paragraph{Prompts and Languages}
After a preliminary analysis of potential prompt formats, our experiments use the following prompts: 
\begin{itemize}
    \item \textbf{Without Context}: The word ``$w_s$'' is translated into $L_t$ as \_\_
    \item \textbf{With Context}: In the sentence ``$c_s$'', the word ``$w_s$'' is translated into $L_t$ as \_\_
\end{itemize}

\noindent We perform experiments with English as the source language and translate into Chinese, French, and Spanish as the target languages.

\paragraph{Models}
We use the GPT-Neo series of LMs with model sizes between 125 million to 20 billion parameters (including the GPT-J model that contains 6B parameters) and the BLOOM series with different model sizes from 560 million to 7.1 billion; all LMs are autoregressive models  trained for next token prediction. We note that BLOOM is explicitly pretrained on all three of our target languages, whereas GPT-NeoX \cite{black2022gptneox20b} is trained as an English LM; however, GPT-NeoX's pretraining corpus contains an estimated $\sim2.6 \%$ of non-English text \cite{gao2020pile}, and prior work has found even small percentages of non-English text can facilitate cross-lingual transfer in English PLMs \cite{blevins2022language}.

\begin{figure*}
    \begin{center}
        \subfloat[]{\includegraphics[width=0.48\textwidth]{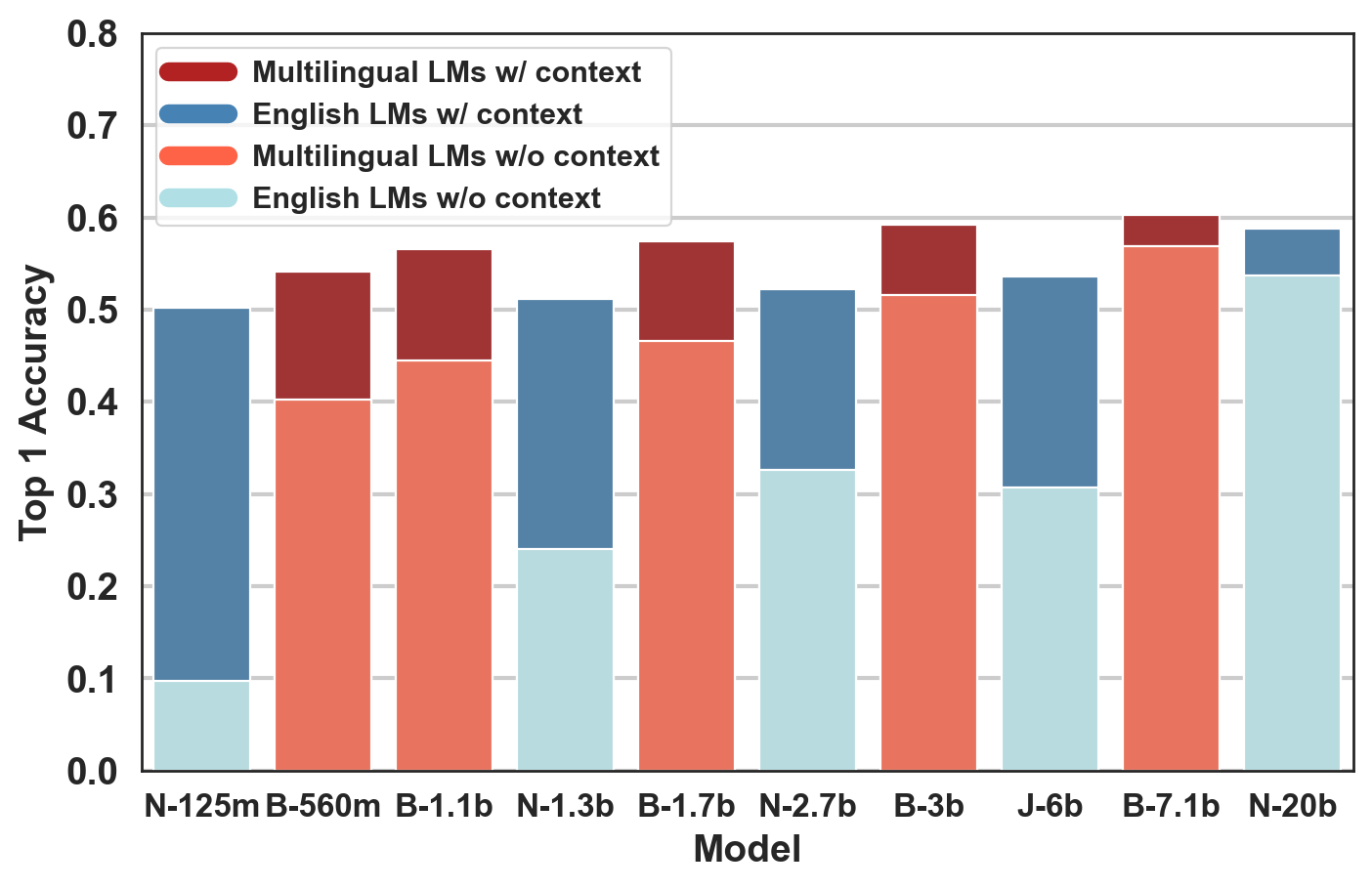}}
        \hfill
        \subfloat[]{\includegraphics[width=0.48\textwidth]{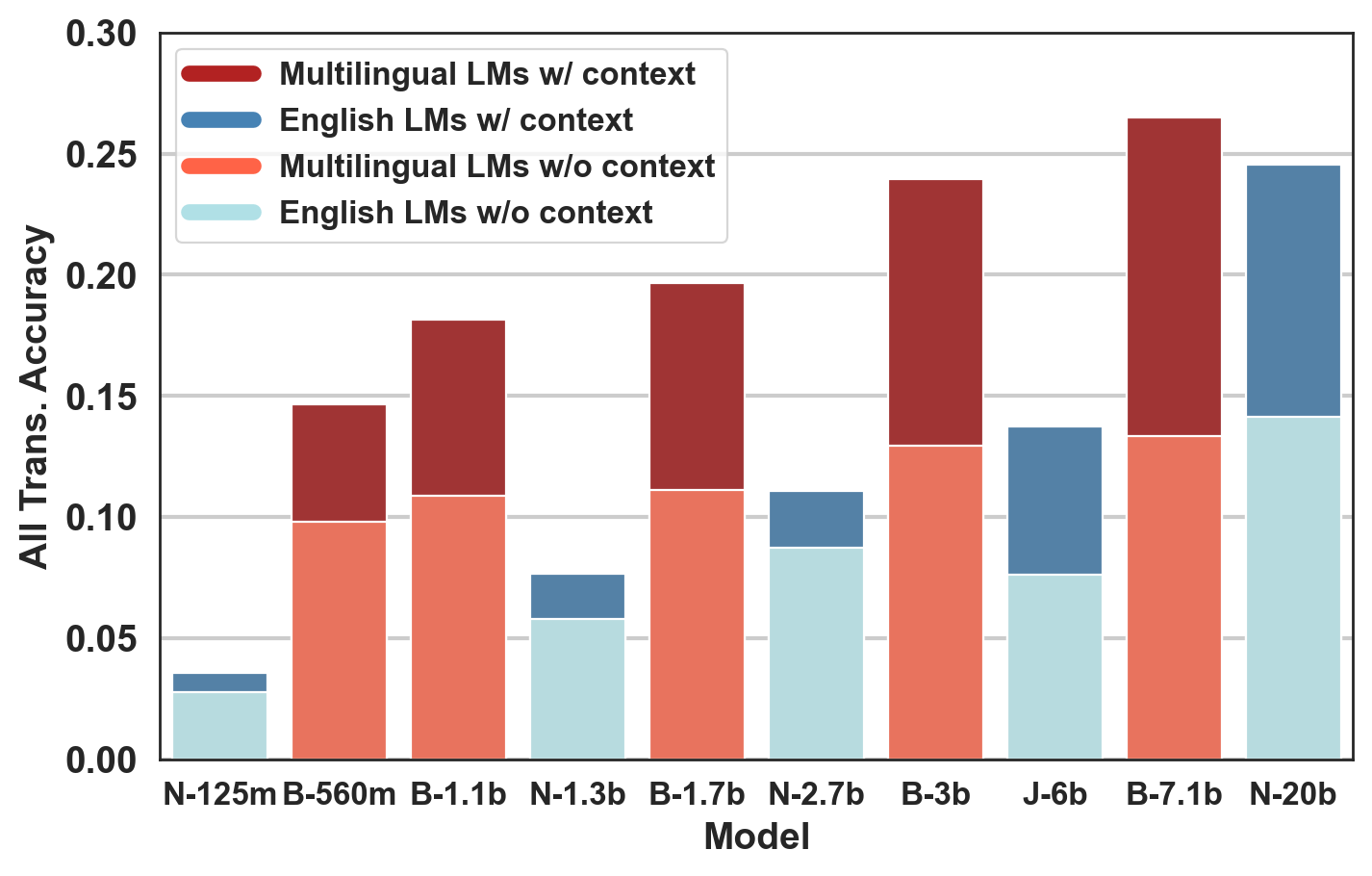}}
    \end{center} 
    \caption{Results of the zero-shot contextual WLT accuracies on GPT and BLOOM family models of different sizes (a) The results of top-1 accuracies across models. (b) The results of all translations accuracies across models. N: GPT-Neo, B: BLOOM, J: GPT-J}
    \label{fig:cwlt-results}
\end{figure*}

\paragraph{Dataset}
We select candidate source words from the English inventory of the XL-WSD dataset \cite{pasini2021xl}. We then filter these into language pair datasets with \textit{<source word, source example context, translations in context>} tuples, where the sense-specific translations and example contexts are obtained from WordNet \cite{miller1995wordnet}.
We include in our dataset the source words where the most common sense (the first sense in WordNet) and at least one other sense meet the following criteria: (a) both senses have non-overlapping sets of translations in the target language and (b) both senses are annotated with example contexts in the source language.
For each sense, we use the translations for the other sense and 50 randomly selected words in the target language as incorrect translations, which are used as negative samples. Due to limited cross-lingual coverage with WordNet, the EN-FR, EN-ES, and EN-ZH experiments include 2448, 2470, and 2084 evaluation examples respectively.

\paragraph{Metrics}
We present three different types of metrics to evaluate the performance of models on the WLT task, with and without context.
\begin{itemize}
    \item \textbf{Accuracy}: We calculate two metrics to measure the accuracy of the models. (1) \textit{top-1 accuracy} measures the percentage of test instances in which the translation with the highest log-likelihood is one of the correct translations for a given sense. (2) \textit{All translations accuracy} measures the percentage of test instances where all $k$ correct translations for that sense are assigned the $k$ highest likelihoods by the model.
    \item \textbf{Negative Log-Likelihoods (NLL)}: We compare the average \textit{negative log-likelihood (NLL)} of all (1) correct and (2) incorrect translations for each sense, as well as (3) the \textit{ratio} of the average NLL of the top-1 correct translations to the average NLL of all incorrect translations for each sense.
    \item \textbf{Error Reduction}: We evaluate the impact of adding context sentences on resolving two types of errors. The first is \textit{disambiguation} errors, where the model produces a valid translation without context that would be an incorrect sense in the additional context; the second is \textit{translation} errors, where the model correctly translates the word in question (based on the context sentence) but produces a mistranslation without context.
\end{itemize}

\subsection{Results}
\label{sec:cwlt-results}
\paragraph{Adding Context Improves Word-Level Translation Accuracy}

Figure \ref{fig:cwlt-results} presents the overall WLT results with and without context, averaged across the three target languages; word-level translation performance improves across all settings with the addition of context.\footnote{The results for each specific target language can be found in the appendix. (Figure \ref{fig:cwlt-chinese} for Chinese; Figure \ref{fig:cwlt-french} for French; Figure \ref{fig:cwlt-spanish} for Spanish)} We also  observe that the performance of both uncontextualized and contextualized word-level translation improves as the model size increases, which corroborates prior findings that larger models better capture cross-lingual information from pretraining \cite[e.g.][]{lin2021few}. 

Our experiments also show that, on average, the multilingual models outperform comparably sized English models in both WLT settings: the multilingual models achieve an average \textit{top-1 accuracy} of 47.94\% in the uncontextualized task and 57.51\% in the contextual task, whereas the English models obtain 30.20\% and 53.2\% in these settings, respectively. However, the performance gap between English and multilingual models narrows when we add sentences that use the word in context. Specifically, the experiments show that the largest English model, GPT-NeoX, performs similarly to the (smaller) multilingual BLOOM models; this suggests that English language models become more effective in leveraging limited cross-lingual knowledge at larger scales. 

Finally, in the setting of \textit{all translations}, we observe that the improvement in performance due to the addition of context is more significant for multilingual models than for English models, leading to the larger performance gaps between these two types of models.

\begin{figure}

    \begin{center} \includegraphics[width=0.48\textwidth, height=5cm]{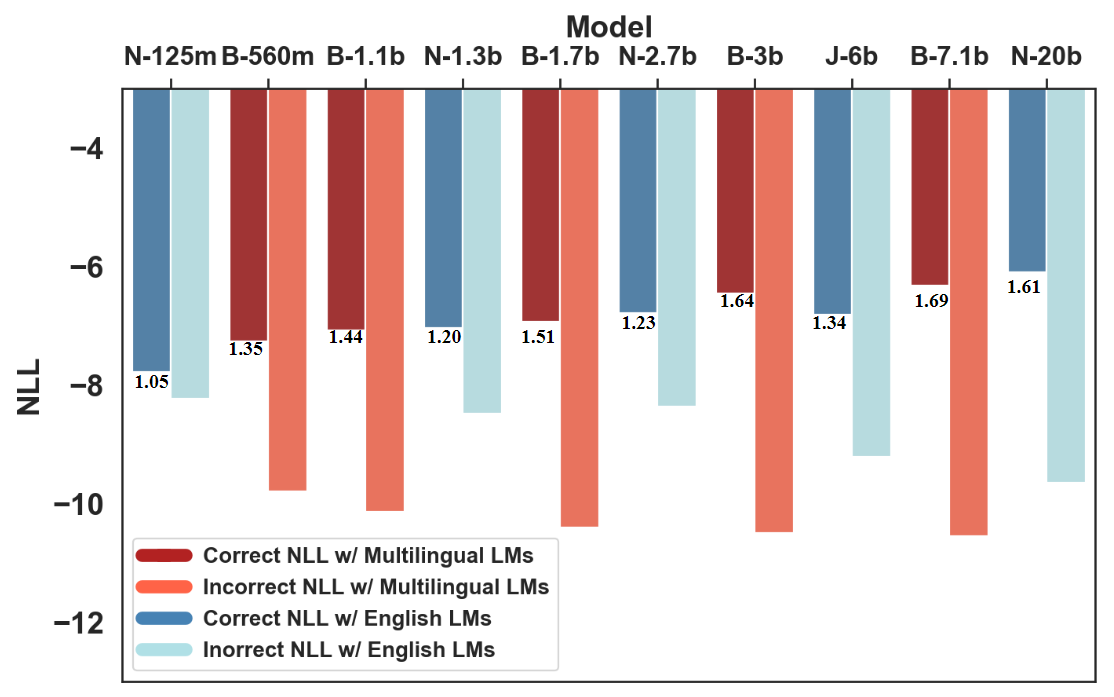}
    \end{center}
    \caption{The average NLL of all correct and incorrect words across models in the contextual WLT analysis (less negative is better). The numbers in the figure represent the ratios of the negative NLLs of incorrect to correct translations (larger is better).}
    \label{fig:cwlt-nll}
\end{figure}

\paragraph{Negative Log-Likelihoods}
We also consider the negative log-likelihoods produced by each model for the top correct translation compared to the incorrect translations (Figure \ref{fig:cwlt-nll}). These results show that the negative log-likelihood (NLL) of the correct translations improves as the model size increases, suggesting that the models become more confident in their predictions in absolute terms. Furthermore, we find that the NLL ratio between correct and incorrect translation words generally increases as the model size improves; the multilingual models also demonstrate better differentiation ability between correct and incorrect translations than English models. Specifically, we observe an average ratio of 1.53 between incorrect and correct translations for multilingual models, compared to 1.28 for English models.

\begin{figure}
    \begin{center} \includegraphics[width=0.48\textwidth]{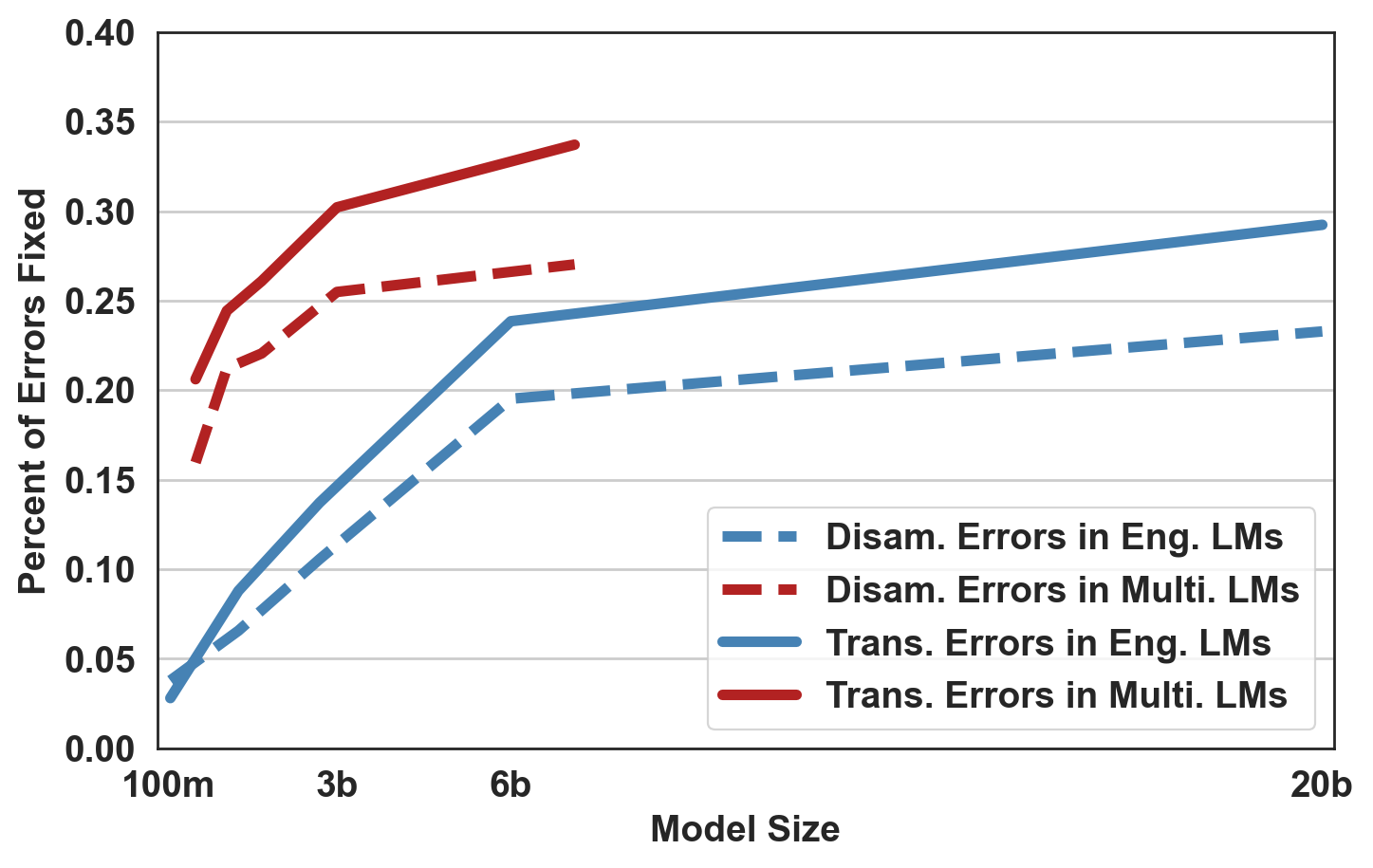}
    \end{center}
    \caption{The impact of adding context to WLT on translation (trans.) and disambiguation (disam.) errors across different model sizes.}
    \label{fig:cwlt-flip}
\end{figure}

\paragraph{Translation Error Reduction with Context}

Finally, we analyze the extent to which adding context sentences fixes errors made by the PLMs in the standard WLT setting (Figure \ref{fig:cwlt-flip}).
Our results show that larger models benefit more than smaller ones from using contextual information to correct translation errors, with a larger percentage of prior errors resolved with the addition of context; this further highlights their ability to better leverage the additional context. In addition, multilingual models fix errors at a higher rate compared to English models  With the addition of context. 

Surprisingly, we also observe that context helps correct complete \textit{translation} errors at higher rates than it does to \textit{disambiguate} the appropriate translation given a context sentence. This generally holds true for both the English and multilingual models and across all model scales, with the smallest English models as an exception (where very few errors of either type are resolved by the addition of context).

\section{Zero-shot Word Sense Disambiguation via C-WLT}

\begin{figure*}
    \begin{center}
        \includegraphics[width=0.85\textwidth]{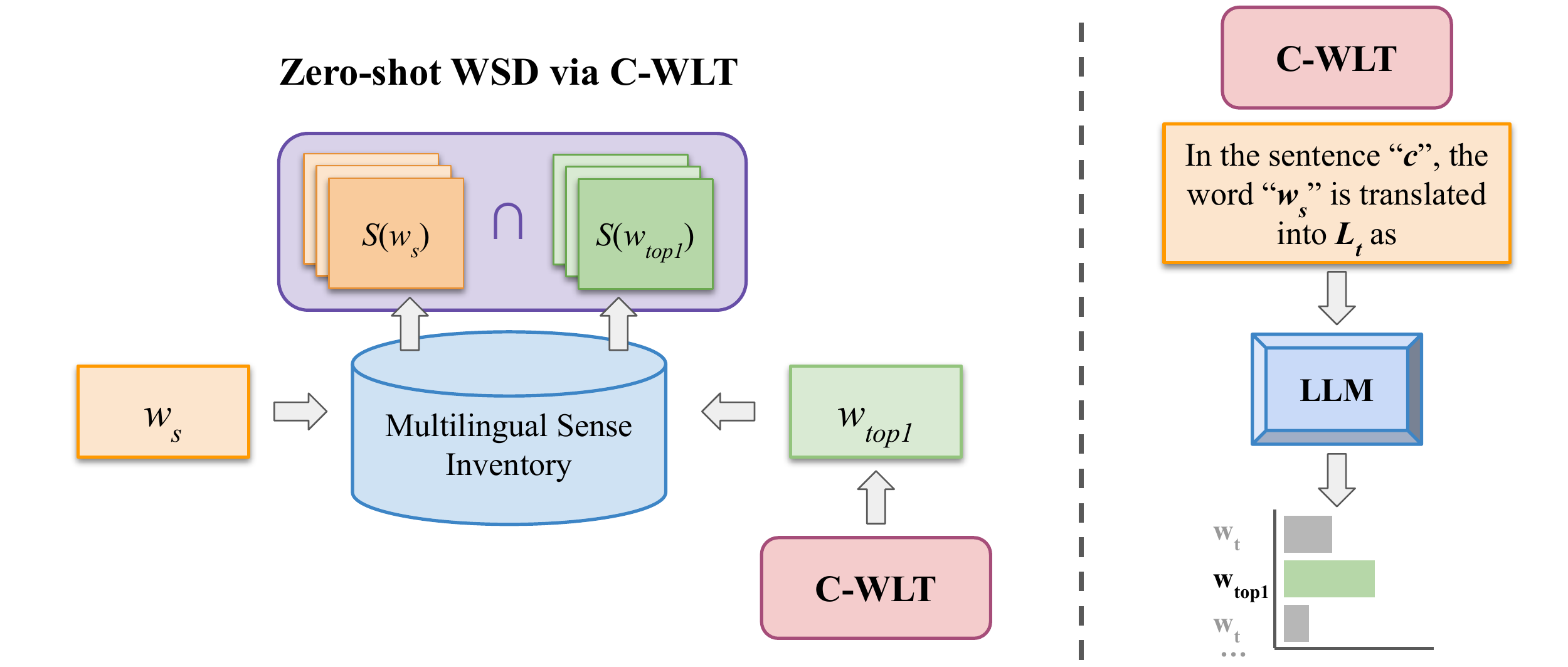}
    \end{center}
    \caption{Overview of the proposed method for multilingual WSD via C-WLT (left) and the prompting setup for C-WLT (right). We translate each ambiguous word $w_s$ in context into a target language $t$ with a PLM and label it with the intersection of its labels and the labels of the translation $w_{top1}$.}
    \label{fig:xl-wsd-method}
\end{figure*}

Building on the intuition from the previous section that contextual word-level translation can differentiate between different meanings of a word in the source language, we apply C-WLT to the task of multilingual word sense disambiguation (Figure \ref{fig:xl-wsd-method}). Specifically, we propose a two-step process wherein we (1) prompt the PLM for C-WLT to translate the word being disambiguated, $w$, in the relevant context and (2) disambiguate $w$ based on the senses of its translation. 

For instance, if we would like to disambiguate the sense of the word ``plant'' as it is used in the context ``The plant sprouted a new leaf'', we would first prompt the PLM to translate ``plant'' into the chosen target language (such as Chinese) using the C-WLT setup from the previous section. We then take the top translation from the PLM (in this case, 
``\zh{植物}'') and obtain its senses from a multilingual word sense ontology. The example is then disambiguated with the set of senses that overlap between the senses of ``plant'' and the senses of ``\zh{植物}''.

\subsection{Method} 
 
The goal of word sense disambiguation (WSD) is to determine the meaning of a word $w$ as it is used in a specific context $c$ and label it with the sense label (or labels) that represents this meaning out of the candidate set of senses associated with that word, $S$. In our proposed approach, \textbf{WSD via C-WLT}, $w$ and $c$ are in a language $L_s$, and word senses are obtained from a multilingual ontology \cite[BabelNet,][]{navigli2010babelnet} and shared across languages.

First, we prompt a PLM with the C-WLT setting to translate $w_s$ based on $c_s$ into the target language $L_t$. 
We then obtain the inventory of all possible translations of $w_s$ into $L_t$ from the multilingual word sense ontology and rank them with the PLM conditioned on the C-WLT prompt. We then label $w_s$ with the set of senses in the intersection of its candidate senses, $S(w_s)$, and those of the top-scoring translation under the PLM, $S(w_{top 1})$.
We note that this means the WSD via C-WLT method assigns a set of labels to $w$ rather than a single sense label like most trained WSD classifiers. 

\paragraph{Ensembling Target Languages}
The described method for WSD via C-WLT obtains potential senses from translating into a single target language. We extend the method to ensemble the senses from a set of target languages $T$, as we hypothesize that senses shared by translations of $w_s$ in multiple typologically diverse languages are more likely to be relevant to the specific context at hand.

Specifically, we consider the multiset of senses for the top translation in every target language: $S(T) = \{S(w_{top1}^t): t \in T \}$. Our target set $S(T)'$ is  the subset of $S(T)$ that contains all senses that share the highest multiplicity (i.e., occur most frequently) in $S(T)$. This means that senses shared by translations of $w_s$ into multiple languages are more likely to be included in $S(T)'$. Similar to the single target language setting, we obtain the final predicted sense set from the intersection of $S(T)'$ and $S(w_s)$.

\subsection{Experimental Setup}

\paragraph{Datasets} 
We evaluate performance with the XL-WSD dataset \cite{pasini2021xl}, which is comprised of 18 languages: Basque, Bulgarian, Catalan, Chinese, Croatian, Danish, Dutch, English, Estonian, French, Galician, German, Hungarian, Italian, Japanese, Korean, Slovenian, and Spanish. We use the BabelNet API 4.0.1 \cite{navigli2010babelnet} as our multilingual word sense ontology to obtain translations and sense inventories of the data.

We consider five target languages for our experiments: English, Chinese, Russian, Spanish, and Finnish; we aim to consider a wide range of typologically diverse languages as targets while maintaining high coverage of the source language examples in the multilingual ontology.\footnote{English covers 100.0\% of the evaluation examples (excluding the EN-coarse set), while Chinese, Spanish, Finnish, and Russian cover 79.0\%, 95.3\%,  99.6\%, and 60.0\%, respectively.} 
In the case where a (non-English) evaluation example does not have at least one corresponding translation in the target language, we back off to the English translation setting as it provides full coverage over all non-English evaluation sets. When evaluating English, we instead back off to the most common sense (MCS) of the word when an example is not covered by the target language(s) in each evaluation setting.

\paragraph{Models}
Picking the three most powerful PLMs from the previous section, we use the BLOOM models \cite{scao2022bloom} with 3 billion parameters and 7.1 billion parameters and the GPT-NeoX model with 20 billion parameters \cite{black2022gptneox20b}. While GPT-NeoX is primarily trained in English, the Bloom models are specifically pretrained on 6 out of the 18 evaluation languages of the XL-WSD dataset.

\paragraph{Baselines}
We consider the Most Common Sense (MCS) method as a baseline, which predicts each word's most common sense according to BabelNet \cite{pasini2021xl}. Additionally, we report the best results from the models introduced to benchmark the XL-WSD dataset in \citet{pasini2021xl} as well as those in \citet{zhang2022disentangled} and \citet{berend-2022-combating}. Prior results are presented as a point of reference for the task scores. However, previous models for the XL-WSD dataset all require supervised training with annotated WSD data, unlike our approach, which is zero-shot and assumes no additional data or finetuning of the PLM.

\paragraph{Evaluation Metrics for WSD via C-WLT}
We consider two automatic metrics for evaluating the performance of the WSD via C-WLT approach. The first is \textit{recall}, or how often the predicted label set contains at least one of the gold annotations for a given example. This metric is obtained from the dataset's evaluation script and is the standard for XL-WSD evaluation; it is often reported as F1 or accuracy in cases where the WSD approach produces a single prediction. 

However, recall overestimates performance in cases where a WSD approach predicts many unrelated sense labels in addition to a correct one. We therefore also calculate the \textit{Jaccard index} between the predicted set and the reference set of sense labels for each example: $\frac{|L_{true} \cap L_{pred}|}{|L_{true} \cup L_{pred}|}$. While the Jaccard index is a better automatic measure of similarity in the setting of sense sets than recall, the metric can underestimate performance in cases where other, closely related senses are also appropriate in the given context but are not included in the reference sense set.\footnote{This type of annotation error is the most common found in an audit of English WSD corpora \cite{maru-etal-2022-nibbling}.}

\begin{table*}[]
    \centering
    \small
    \begin{tabular}{c | r r | r r r|| r r r}
        \toprule
       \multirow{2}{*}{\textbf{Language}}  & \multirow{2}{*}{\textbf{MCS}} & \multirow{2}{*}{\textbf{Prior Work$^{*}$}} & \multicolumn{3}{c||}{\textbf{Recall}} & \multicolumn{3}{c}{\textbf{Jaccard Index}} \\
       & & & \textbf{NeoX} & \textbf{B-3B} & \textbf{B-7.1B} & \textbf{NeoX} & \textbf{B-3B} & \textbf{B-7.1B} \\
       \hline
        Basque & 32.72 & 51.71 (b) & 47.85 & \underline{52.53} & \textbf{\underline{54.31}} & 37.20 & \underline{41.04} & \textbf{\underline{42.95}}\\
        Bulgarian & 58.16 & 73.60 (a) & \textbf{75.51} & 71.56 & 72.05 & \textbf{66.28} & 63.32 & 63.78\\
        Catalan & 27.17 & \textbf{57.47} (b) & 55.73 & \underline{55.83} & \underline{56.40} &  39.44 &  \underline{40.41} & \underline{\textbf{40.85}}\\
        Chinese &  29.62 & 57.05 (b) & \textbf{61.03} & \underline{60.64} & \underline{58.87} & \textbf{46.86} & \underline{46.78} & \underline{46.26}\\
        Croatian & 62.88 & 74.40 (b) & \textbf{77.01} & 74.85 & 74.82 & \textbf{70.00} & 68.53 & 68.46\\
        Danish & 64.33 & 81.80 (c) & \textbf{81.86} & 76.76 & 77.38 & \textbf{73.50} & 69.69 & 70.32\\
        Dutch &  44.61 & 61.95 (b) & \textbf{66.25} & 61.89 & 63.46 & \textbf{55.72} & 52.07 & 53.33 \\
        English$^{\dagger}$ & 63.37 & \textbf{76.77} (a) & 72.61 & \underline{72.15} & \underline{73.20} & 60.56 & \underline{60.13} & \textbf{\underline{61.39}}\\
        Estonian & 46.87 & 68.88 (b) & \textbf{70.24} & 65.58 & 65.88 & \textbf{61.72} & 58.94 & 58.80\\
        French & 59.31 & \textbf{83.88} (a) & 76.04 & \underline{76.47} & \underline{78.02} & 64.67 & \underline{65.62} & \textbf{\underline{68.00}}\\
        Galician & 60.85 & 67.3(b) & 74.15 & 74.63 & \textbf{74.82} & 60.47 & \textbf{61.06} & 60.84\\
        German & 75.99 & \textbf{84.69} (b)& 81.45 & 78.31 & 81.57 & \textbf{74.40} & 71.60 & 74.02\\
        Hungarian & 47.29 & \textbf{76.4} (c) & 75.52 & 71.56 & 72.04 & \textbf{66.28} & 63.32 & 63.77\\
        Italian &  52.77 & \textbf{77.8} (c) & 76.63 & 74.50 & 74.58 & \textbf{57.91} & 57.62 & 57.63\\
        Japanese &  48.71 & 67.47 (c) & \textbf{71.63} & 70.78 & 71.38 & 57.56 & 57.38 & 55.72\\
        Korean & 52.48 & \textbf{68.2} (c) & 66.39 & 67.52 & 67.73 & 60.95 & 61.01 & 61.46\\
        Slovenian & 36.71 & \textbf{68.36} (a) & 53.12 & 46.21 & 47.93 & \textbf{40.32} & 33.36 & 37.05\\
        Spanish & 55.65 & 76.93 (b) & 75.42 & \underline{75.53} & \textbf{\underline{77.66}} & 55.58 & \underline{56.50} & \textbf{\underline{58.36}}\\
        \hline
        Avg. & 49.31 & -- & 70.35 & 68.62 & 69.45 & 58.59 & 57.42 & 58.24 \\ 
        \toprule
    \end{tabular}
    \caption{Zero-shot Recall and Jaccard Index for multilingual WSD on the XL-WSD dataset in the best-ensembled setting. Results for languages on which Bloom was pre-trained are \underline{underlined}. $^{*}$Prior work numbers are drawn from the best results reported in (a) \citet{pasini2021xl}, (b) \citet{berend-2022-combating}, and (c) \citet{zhang2022disentangled}; note that prior approaches are \textit{not} zero-shot as they require finetuning on labeled WSD data. $^{\dagger}$For the 1512 (out of 8062) English examples with coverage issues, we used MCS as predictions.} 
    \label{tab:wsd-main-results}
\end{table*}

\section{Multilingual WSD Results and Analysis}
\label{sec:xl-wsd}
We first present the performance of our method for multilingual WSD on the two automatic metrics, recall and Jaccard index, and compare this approach to prior work on this task (Section \ref{sec:main-wsd-results}). We then consider the effect of ablating different modeling choices on our method (such as the choice of target language for C-WLT and prompt language; Section \ref{sec:ablations}), and we analyze the outputs and types of errors the approach produces more closely (Section \ref{sec:error-analysis}). 

\subsection{Results}
\label{sec:main-wsd-results}

The multilingual WSD results are summarized in Table \ref{tab:wsd-main-results}. In our experiments, we found that the best setting for achieving a balance between recall and Jaccard Index was to ensemble English, Chinese, and Russian as the target languages with English prompts (Table \ref{tab:diff_targets_neoX}). The results show that our approach achieves higher recall compared to the prior works in 11 out of the 18 source languages, despite the fact that our method is performed zero-shot from a pretrained language model. If considered as an upper bound measure on performance, this result shows that translation-based approaches for WSD can identify the correct sense label(s) as well or better than supervised methods.

We also find that despite being primarily pretrained on English, GPT-NeoX (20B) achieves higher recall and Jaccard index scores than Bloom-7.1 on 10 source languages; most settings where the multilingual model performs better are on its pretraining languages, with little generalization to other languages. Finally, despite the Jaccard index scoring lower (by definition) than recall, we see similar performance trends across languages and models between recall and the Jaccard index in this ensemble setting.


\subsection{Modeling Ablations} 
\label{sec:ablations}

\paragraph{Different Target Languages} 
We consider five different target languages: English, Chinese, Russian, Finnish, and Spanish. In addition to the five individual target language settings, we experiment with all combinations of joint target language settings (Table \ref{tab:diff_targets_neoX}).\footnote{We report the Bloom results in Table \ref{tab:diff_targets_bloom} in the appendix; we observe similar tradeoffs when using those models.} We also calculate the \textit{delta} increase in the sense prediction rates, normalized by the number of senses for each example. We compare the standard classification setting of predicting a single label per WSD example, and the number of labels predicted by each target language setting: $\frac{1}{n}\sum_{i=0}^{n} \frac{|\hat{S}_i|}{|S_i|} - \frac{1}{n}\sum_{i=0}^{n} \frac{1}{|S_i|}$ where $S_i$ is the candidate sense set for the $i$th evaluation example and $\hat{S}_i$ is the set of senses predicted by our approach.

Our ablations indicate a tradeoff between the Jaccard index and recall metrics. For example, our approach achieves the highest recall performance using Spanish as the sole target language, but the resulting Jaccard index is worse than any other target setting we test. This behavior is likely because target languages more similar to the source (such as Spanish, which is closely related to many of the Western European source languages in the XL-WSD dataset) return a larger set of predicted senses, which in turn improves recall but at the expense of set similarity with the gold labels. This hypothesis is corroborated by the high delta increase of 20\% in the predicted set size of the Spanish setting over the standard single-label predicted setting. 

However, this undesirable behavior is mitigated by using less similar target languages and by ensembling a diverse set of languages. In our best setting of ensembling English, Chinese, and Russian we find that the delta increase in the predicted set size  is only 6.7\%, while the Jaccard index increases by $\sim$6 points over Spanish. Furthermore, this ensembled setting still often outperforms prior approaches on recall.

 \begin{table}[b!]
    \centering
    \begin{tabular}{c|c| c||c}
    
         \textbf{Target Lang.} & \textbf{Recall} & \textbf{Jaccard} & \textbf{Delta}$^{*}$ \\
        \hline
        Spanish & 74.23 & 52.94 & 20.0\\
        English & 67.16 & 53.37 & 11.7\\
        Finnish & 66.35 & 54.28 & 12.9\\
        Russian & 67.42 & 55.08 & 10.2\\
        Chinese & 70.84 & 57.77 & 9.6 \\
        \hline
        Best Setting & 70.35 & 58.59 & 8.7 \\
        All 5 Joint$^{\dagger}$ & 66.60 & 57.50 & 6.7\\
    \end{tabular}
    \caption{The average Recall and Jaccard Index (\%) for the different target language settings of the GPT-NeoX model, as well as the delta(*) increase in sense label prediction rates. $^{\dagger}$``All 5 joint'' refers to the setting of using all five target languages above, whereas the ``best setting'' ensembles English, Chinese, and Russian.}
    \label{tab:diff_targets_neoX} 
\end{table}

\begin{figure}
    \begin{center} \includegraphics[width=0.5\textwidth]{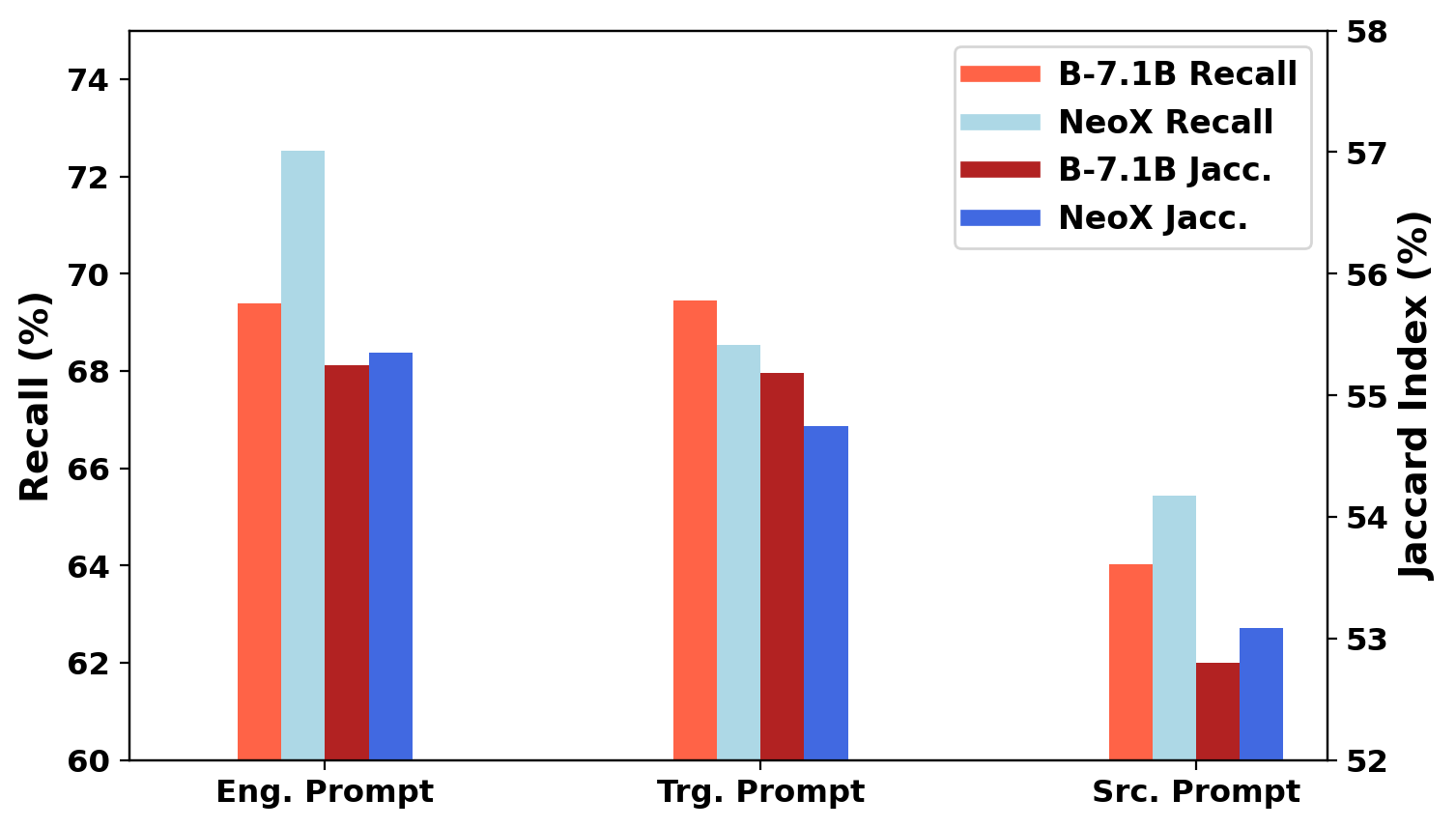}
    \end{center}
    \caption{The results of performance differences by using prompts in different languages. The blue and red bars represent the results of GPT-NeoX and BLOOM-7.1b, respectively.}
    \label{fig:wsd-perf-diff-prompt-mean}
\end{figure}

\begin{figure}
    \begin{center} \includegraphics[width=0.5\textwidth]{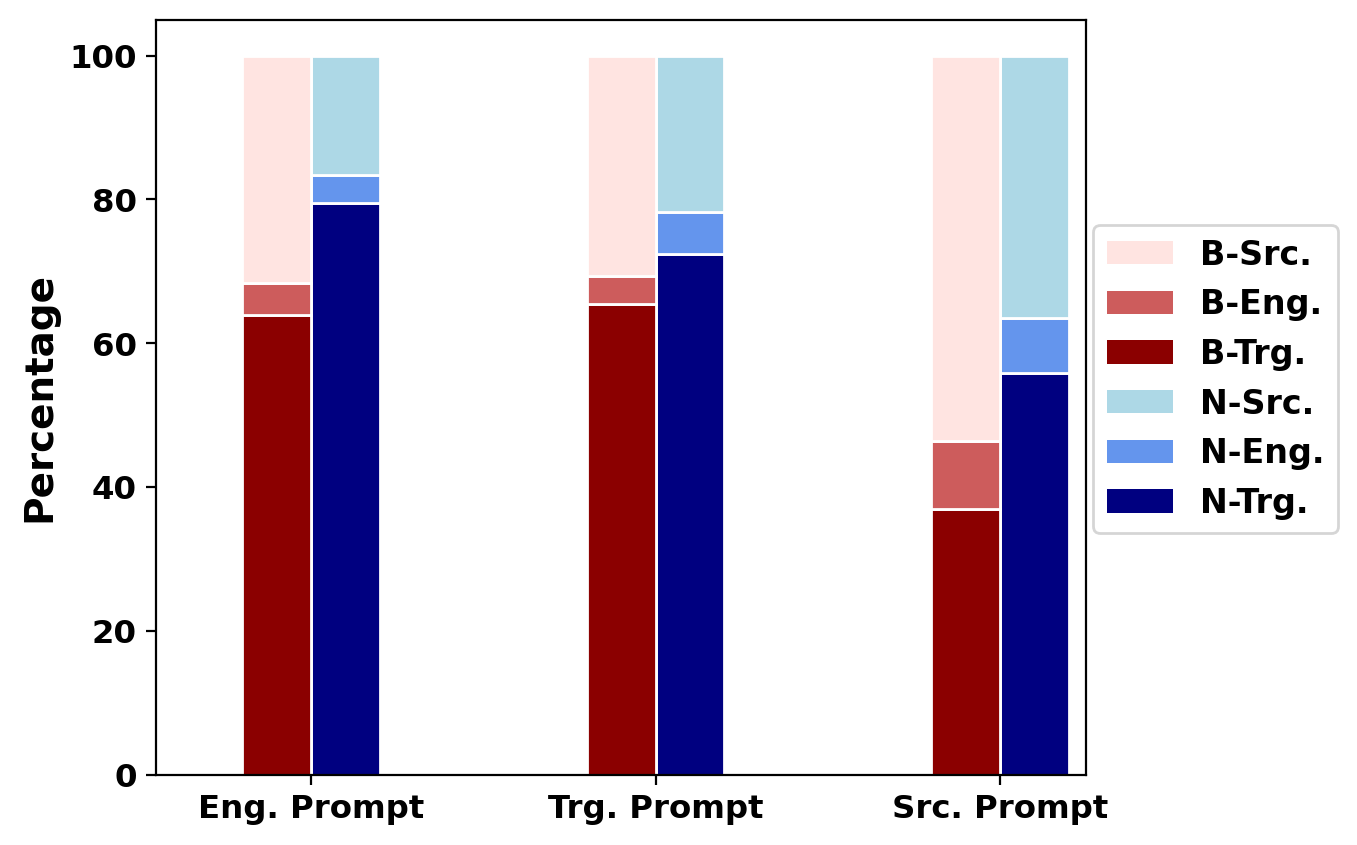}
    \end{center}
    \caption{The proportions of top 1 predictions in different languages by using prompts in different languages. A larger darkest area indicates better performance.}
    \label{fig:wsd-rank-diff-prompt-mean}
\end{figure}

\paragraph{Prompts in Different Languages} 
\label{tab:prompt-diff-lang}
We also consider the effect of prompt language on the WSD via C-WLT method by ablating English prompts, prompts in the source language, and prompts in the target language for C-WLT. The English, Chinese, French, and Spanish prompts were obtained from or verified by native speakers; prompts in other languages were obtained directly from Google Translate. In this study, we use a subset of the evaluation languages, Spanish and Chinese, as our target languages and evaluate based on (a) the overall performance of the method in the prompt language (Figure \ref{fig:wsd-perf-diff-prompt-mean}) and (b) the language of the top-scoring prediction for each prompt language setting, out of the union of the candidate word sets from the prompt, source, and target languages (Figure  \ref{fig:wsd-rank-diff-prompt-mean}). 

We observe that prompts in English and target languages outperform prompts in the source languages, with English prompts generally performing the best (though the target language prompts are comparable to English in Bloom).   We also find that the non-English prompts are more likely to produce a top-1 prediction in the wrong (not target) language. This is particularly true in the case of source language prompts; along with the observed performance decrease, this suggests that prompting the model to generate a label in a different language than the prompt itself is difficult -- unless the prompt language is English. Moreover, our results show that the multilingual LM (BLOOM-7.1b) is more prone to predicting words in the wrong languages than the English LM (GPT-NeoX).

\subsection{Error Analysis}
\label{sec:error-analysis}

\paragraph{Effect of Sense Frequency on Performance}
Supervised WSD classifiers often learn to predict more commonly seen senses in the training data, which leads to stronger performance on examples of the most common sense (MCS) of words than the less common senses (LCS) \cite{maru-etal-2022-nibbling}. We test whether this behavior holds with the unsupervised WSD via C-WLT approach by evaluating performance on examples where the gold sense is the MCS of the word and those  annotated with an LCS separately (Table \ref{tab:mcs-lcs-results}).

The results show that the gap between MCS and LCS performance is relatively large for both metrics: we observe an average difference of 28.7 and 36.3 between MCS and LCS examples for recall and Jaccard index, respectively.  We also find that the size of this performance gap is consistent between the GPT-NeoX and Bloom-7.1B models. We hypothesize that this performance gap stems from unbalanced latent sense supervision in the pretraining data that is due to the natural Zipfian distribution of senses in language \cite{kilgarriff2004dominant}. This finding then highlights that even zero-shot methods extrapolating from the pretraining signals are still vulnerable to unbalanced data.

\begin{table}[b!]
    \centering
    \begin{tabular}{c|c c|c c}
        \multirow{2}{*}{\textbf{Label Set}} & \multicolumn{2}{c|}{\textbf{Recall}} & \multicolumn{2}{c}{\textbf{Jaccard}} \\
        & \textbf{NeoX} & \textbf{B-7.1B} & \textbf{NeoX}  & \textbf{B-7.1B} \\
        \hline
        Orig. & 63.78 & 57.74 & 52.01 & 50.98 \\
        Annot. & \textbf{74.01} & \textbf{74.54} & \textbf{54.29} & \textbf{52.73}
    \end{tabular}
    \caption{The results for the subset of the Chinese evaluation set that was re-annotated in comparison to the original labels of the dataset.}
    \label{table:wsd-annoated}
\end{table}

\paragraph{Manual Precision Analysis}

Based on our observation, the gold annotations in the test sets across all 18 languages mostly consist of one label (and occasionally two). This leads us to hypothesize that there may be other closely related senses that are suitable in the given context but not included in the reference sense set. To investigate this further, we have three native language speakers manually re-annotate 392 examples in the Chinese test set. Interestingly, our analysis finds that 172 examples (or 44\%) have additional, closely related senses that are not included in the original annotations. 

For example, consider the sentence: \begin{CJK}{UTF8}{gbsn}“广播还没说完，各班的同学早已纷纷冲出教室。”\end{CJK} In the XL-WSD dataset, the word \begin{CJK}{UTF8}{gbsn}“广播” \end{CJK} is labeled with the definition, \textit{"Be broadcast"}. 
However, our annotation adds an additional sense with the definition, \textit{"Broadcast over the airwaves, as in radio or television"} into the reference set. 

The results on the subset of the evaluation set we consider show that, unsurprisingly, both the recall and Jaccard index of GPT-NeoX and BLOOM-7.1b improve over the original annotations (Table \ref{table:wsd-annoated}). Therefore, we conclude that the fine-grained annotation errors negatively impact our results, and the additional labels we discover in the annotation indicate that the reference senses may not contain full coverage of relevant senses for many examples. This suggests that future research on multilingual WSD should consider the choice of reference sense sets to ensure that they reflect the full range of senses relevant to a given context, as prior work has done for English \cite{maru-etal-2022-nibbling}.

\section{Discussion and Related Work}
We present a new study of prompting language models for word-level translation \cite{gonen2020greek, li2022improving} in the new setting of contextual word-level translation (C-WLT) to evaluate how well models produce context-sensitive lexical translations. Other related work has instead tested the efficacy of prompting multilingual PLMs for sentence-level translation, such as \citet{lin2021few, vilar2022prompting}. Notably, \citet{bawden2023investigating} observe the issue of incorrect language prediction with multilingual PLMs that we also find (Table \ref{tab:prompt-diff-lang}), particularly for zero-shot prompting. 

We then use the C-WLT setup to perform zero-shot multilingual WSD. This approach builds on \citet{pasini2021xl}, which presents a standardized dataset for WSD in many languages and highlights the role of multilingual language models in addressing the knowledge acquisition bottleneck problem in WSD. Other works on this direction have proposed different finetuning improvements to better perform WSD cross-lingually \cite{zhang2022disentangled, berend-2022-combating}. Unlike prior approaches for the XL-WSD task, our method is zero-shot and does not depend on annotated data; instead, its performance relies on the translation abilities of the PLM.

Additionally, our proposed method is, to the best of our knowledge, the first attempt to apply large-scale autoregressive PLMs to the task of word sense classification via in-context learning. Other work on prompting for word sense has instead framed WSD as a binary classification task in which the model predicts whether the target word in two given contexts with the same target word is used in the same sense in both \cite{pilehvar2019wic, raganato-etal-2020-xl}. 

More generally, WSD is closely related to and motivated by machine translation. A commonly proposed use case of WSD systems is to improve the translation of ambiguous words in MT; as such, multiple methods to incorporate word sense information (such as sense embeddings) into NMT systems have been proposed \cite[e.g.,][]{liu2018handling, campolungo2022reducing}. Furthermore, word sense knowledge has been used to evaluate NMT systems \cite{campolungo2022dibimt}. Prior work has also leveraged MT systems and data to improve an underlying WSD classifier \cite{luan2020improving} and automatically annotate WSD data \cite{diab2002unsupervised, apidianaki2015limsi, hauer2021semi, barba2021mulan}. We build on this latter line of work's intuition to extrapolate word senses from the translations of ambiguous words in context.

\section{Conclusion}
In this work, we examine the ability of pretrained language models to utilize contextual information in cross-lingual settings. Specifically, we propose contextual word-level translation (C-WLT) and test different PLMs' ability to improve lexical translations in context. We then propose a zero-shot technique for multilingual WSD that uses C-WLT as a component and demonstrate its effectiveness on 18 languages, including those with scarce resources or not included in the PLM's pretraining. 

The performance of WSD via C-WLT relies on the relationship between pretraining languages, model size, and the choice of the target language: smaller multilingual PLMs are more effective for languages on which they have been pretrained but are more sensitive to design choices, lacking the broad applicability of their larger English counterparts. Future research examining these interactions and their tradeoffs more closely is vital for improving zero-shot WSD approaches and building better cross-lingual applications of PLMs in general.

\section*{Limitations}
We recognize several limitations that influence C-WLT and our proposed approach for WSD. First, the WSD via C-WLT method is dependent on the composition of the multilingual word sense ontology we use to obtain cross-lingual word senses and translations. Lower coverage in the chosen target language will hinder the method's performance: we see this empirically in the case of English as an evaluation language, as no target language setting (including ensembling) fully covers English, which requires us to back off the MCS of each word.  

Similarly, the translation capability of PLMs, particularly for low-resource languages, may limit the effectiveness of both C-WLT and our WSD approach that relies on it. While we first present a study of the efficacy of C-WLT before incorporating it into our WSD method, due to data limitations (i.e., constructing a C-WLT data for each language pair that contains examples covering multiple senses of many different target words), we examine three high-resource language pairs. However, as better cross-lingual PLMs are developed, they can be directly integrated into our proposed approach to improve WSD for these languages.

Finally, our approach is not well-suited for distinguishing between very fine-grained differences in word sense. While our small-scale manual precision analysis (Section \ref{sec:error-analysis}) suggests that at least some WSD evaluation sets are not annotated with complete coverage of all relevant senses -- leading to an underestimate of our approach's performance -- the ability to differentiate between closely related senses precisely remains a hurdle for the WSD via C-WLT method, and addressing this issue in the future will further improve its applicability.

\section*{Acknowledgements}
We thank the human annotators for the manual precision analysis of the Chinese XL-WSD evaluation set. We also thank Hila Gonen for her helpful comments and discussion on this work.

\bibliography{anthology,custom}
\bibliographystyle{acl_natbib}

\appendix
\label{sec:appendix}

\section{Additional Experimental Details}
\label{app:exp-details}
We present the full set of C-WLT prompts for all 18 evaluation languages from Section \ref{sec:xl-wsd} in Table \ref{tab:wsd_tlang_temp}. We note that for the templates with a [target word], the context prior to [target word] is fed into the PLM as the prompt, and candidates in a target language are concatenated with the part after [target word] to calculate the final score of each potential translation.


\begin{table*}[]
\resizebox{\textwidth}{!}{
    \begin{tabular}{|c|c|}
        \hline
         \textbf{Lang.} & \textbf{Prompt Template} \\
         \hline
         English & In the sentence ``<sentence>'', the word ``<source word>'' is translated into <target langage> as  \\
         Spanish &  En la oración ``<sentence>'', la palabra ``<source word>'' se traduce al <target lang> como  \\
         Chinese & \begin{CJK}{UTF8}{gbsn}在``<sentence>''这句话中, ``<source word>''这个词翻译成<target language>为 \end{CJK}\\
        Catalan & A la frase ``<sentence>'', la paraula ``<source word>'' es tradueix <target lang> com a  \\
        Basque & ``<sentence>'' esaldian, ``<source word>'' <target lang> [target word] gisa itzultzen da \\
        German & In dem Satz „<sentence>“ bedeutet das Wort „<source word>“ ins <target lang> als \\
        Estonian & Lauses ``<sentence>'' tõlgitakse sõna ``<source word>'' <target lang> keelde kui \\
        French & Dans la phrase ``<sentence>'', le mot ``<source word>'' se traduit en <target lang> par \\ 
        Bulgarian & \selectlanguage{bulgarian} В изречението „<sentence>“ думата „<source word>“ се превежда на <target lang> като \\
        Croatian & U rečenici ``<sentence>'', riječ ``<source word>'' prevedena je na <target lang> kao \\
        Danish & I sætningen ``<sentence>'' oversættes ordet ``<source word>'' til <target lang> som `` \\
        Dutch & In de zin ``<sentence>'' vertaalt het woord ``<source word>'' zich in het <target lang> als `` \\
        Galician &  Na frase ``<sentence>'', a palabra ``<source word>'' tradúcese ao <target lang> como \\
        Hungarian & A ``<sentence>'' mondatban fordítsa le a ``<source word>'' szót <target lang> \\
        Italian & Nella frase ``<sentence>'', la parola ``<source word>'' si traduce in <target lang> come \\
        Japanese & \begin{CJK}{UTF8}{min}「<sentence>」という文で、「<source word>」という単語は<target lang>に訳すと [target word] となります \end{CJK}\\
        Slovenian & V stavku ``<sentence>'' se beseda ``<source word>'' v <target lang> prevede kot \\
        Korean & \begin{CJK}{UTF8}{}\CJKfamily{mj}``<sentence>''이라는 문장에서 ``<source word>''이라는 단어는 <target lang> [target word]로 번역됩니다 \end{CJK} \\
         \hline
    \end{tabular}
    }
    \caption{C-WLT templates we used in the experiment for different prompt languages.}
    \label{tab:wsd_tlang_temp}
\end{table*}

\section{Full Experimental Results}
\begin{figure*}

    \begin{tabular}{cc}
        \includegraphics[width=0.48\textwidth]{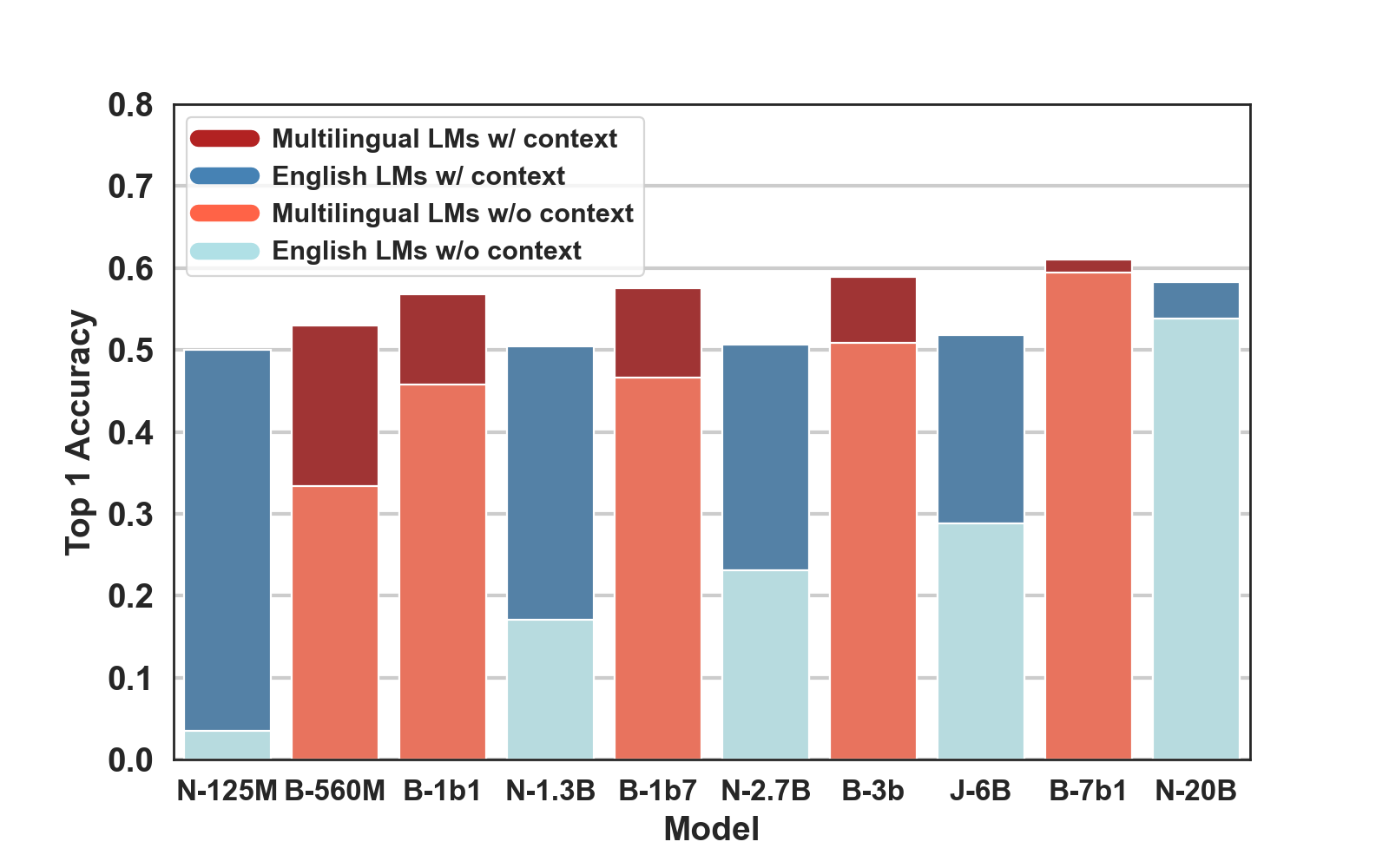} &   
        \includegraphics[width=0.48\textwidth]{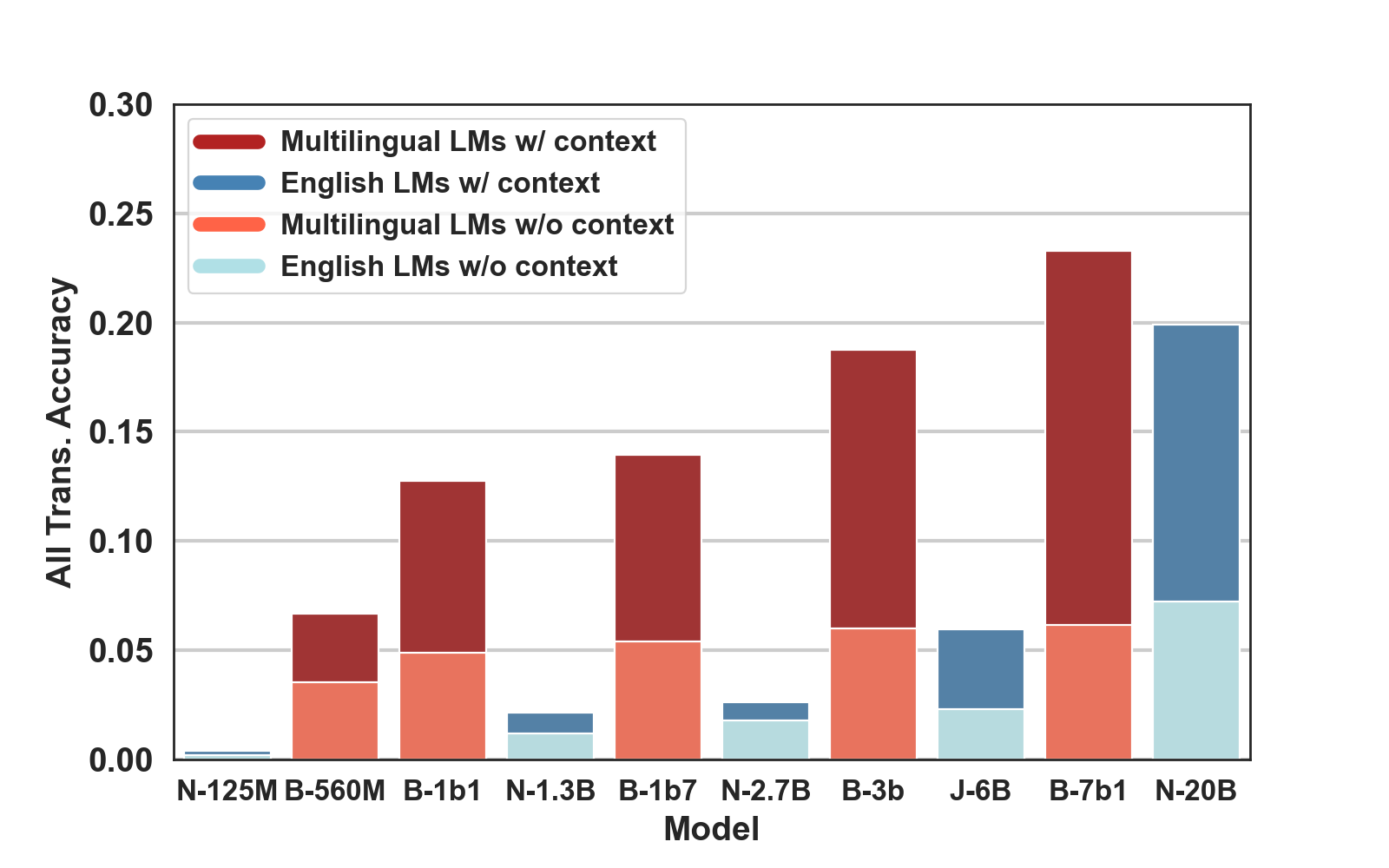} \\
        \includegraphics[width=0.48\textwidth]{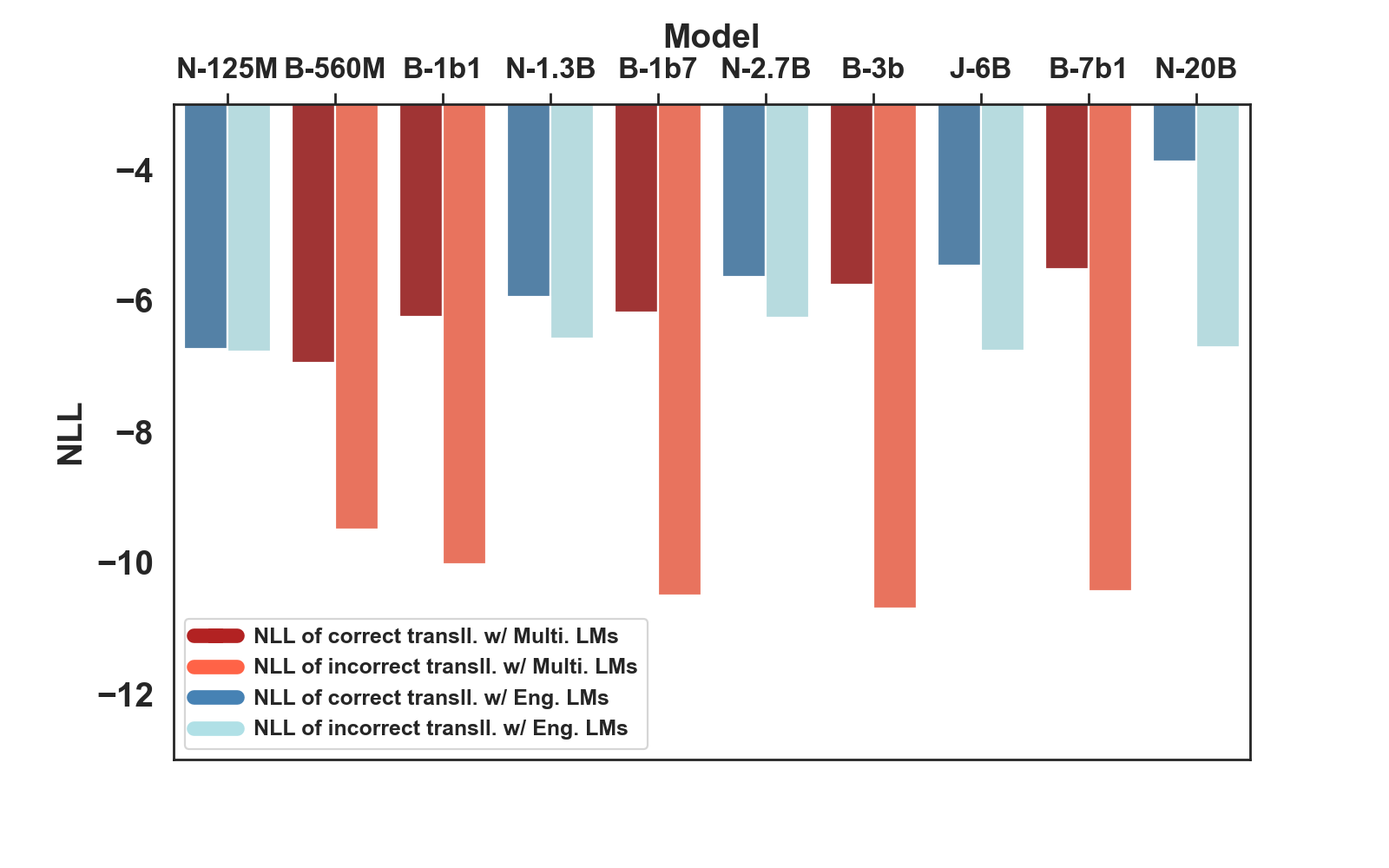} &   
        \includegraphics[width=0.48\textwidth]{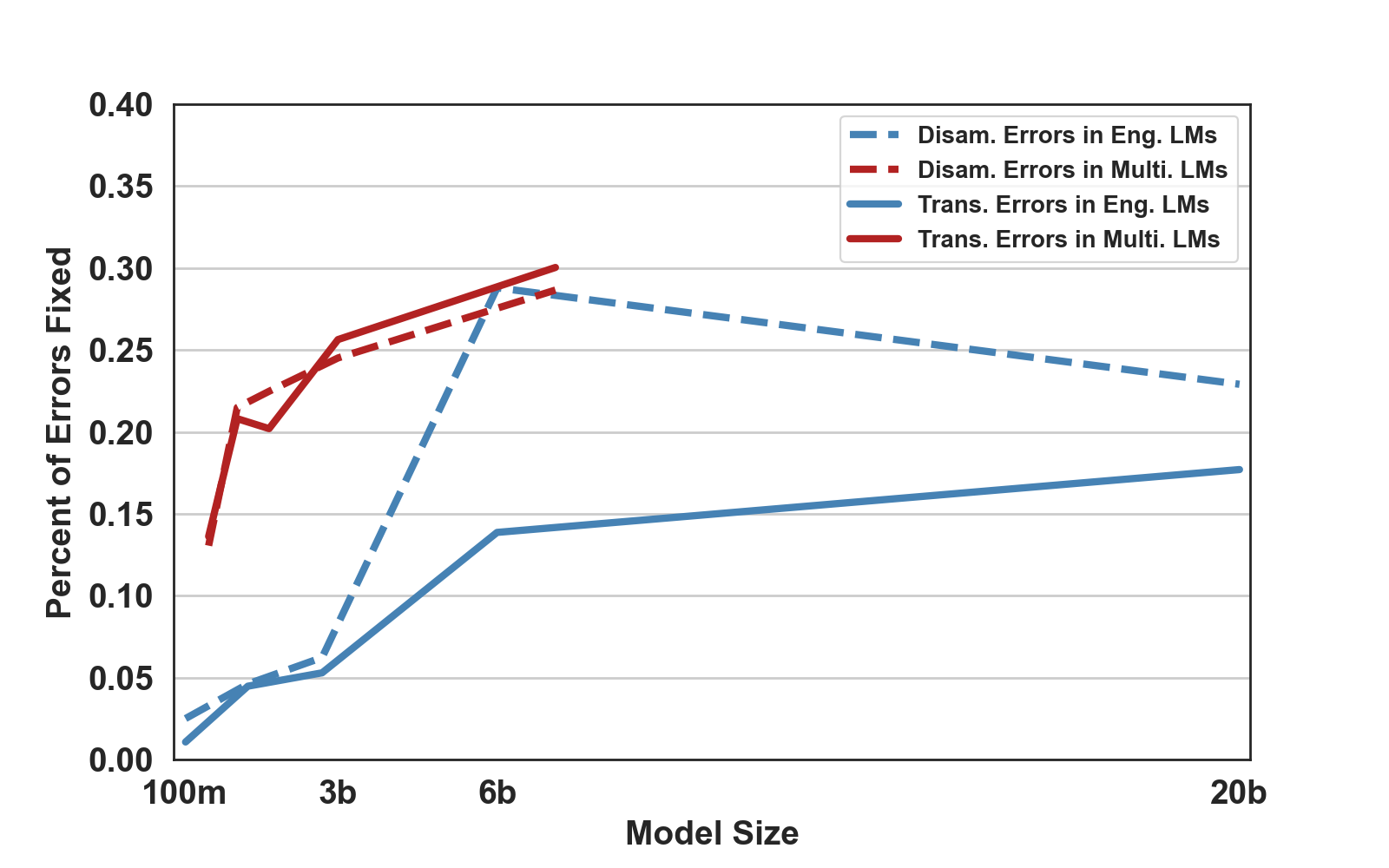} \\
    \end{tabular}
    \caption{C-WLT results for Chinese}
    \label{fig:cwlt-chinese}
\end{figure*}

\begin{figure*}
        \begin{tabular}{cc}
        \includegraphics[width=0.48\textwidth]{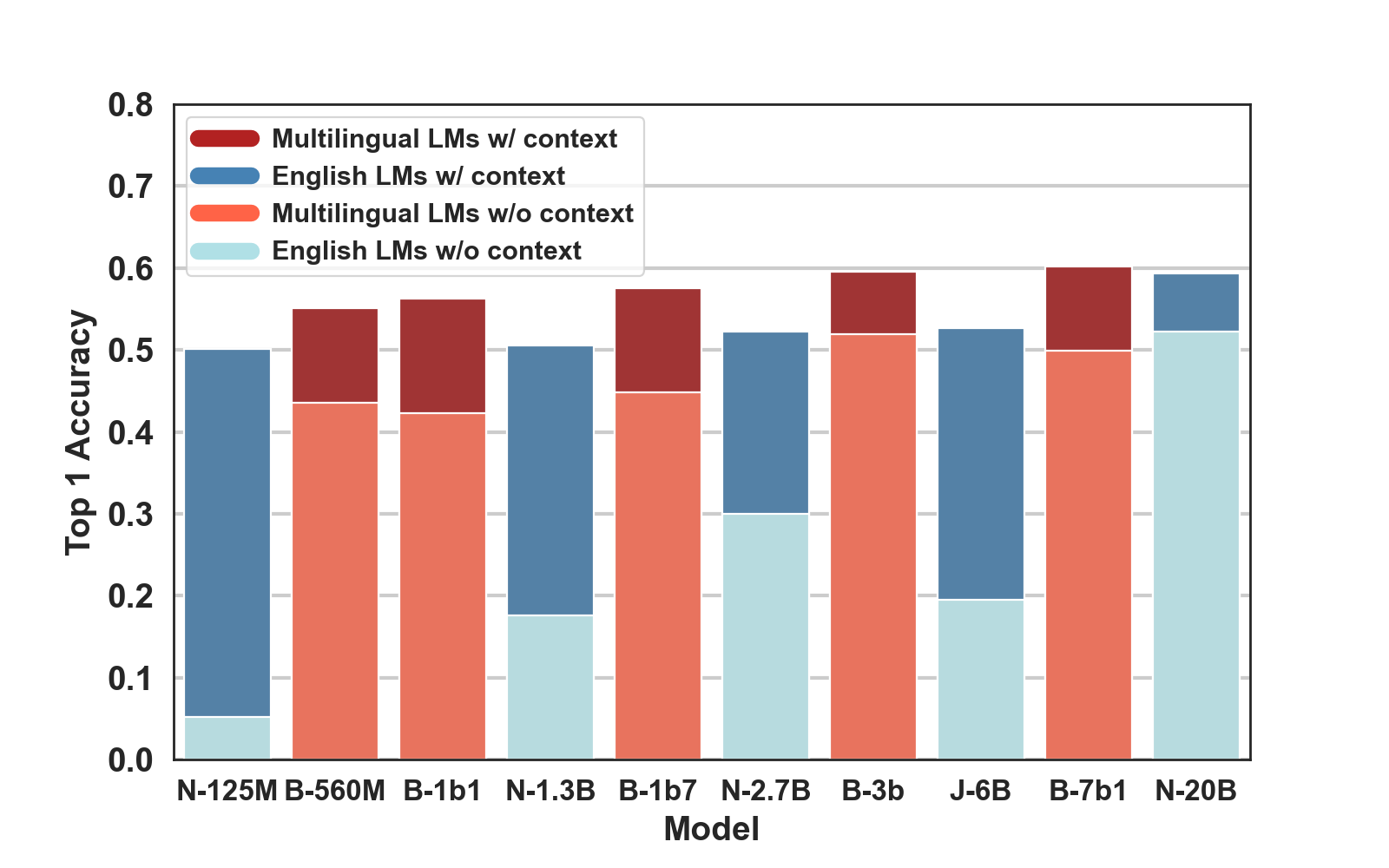} &   
        \includegraphics[width=0.48\textwidth]{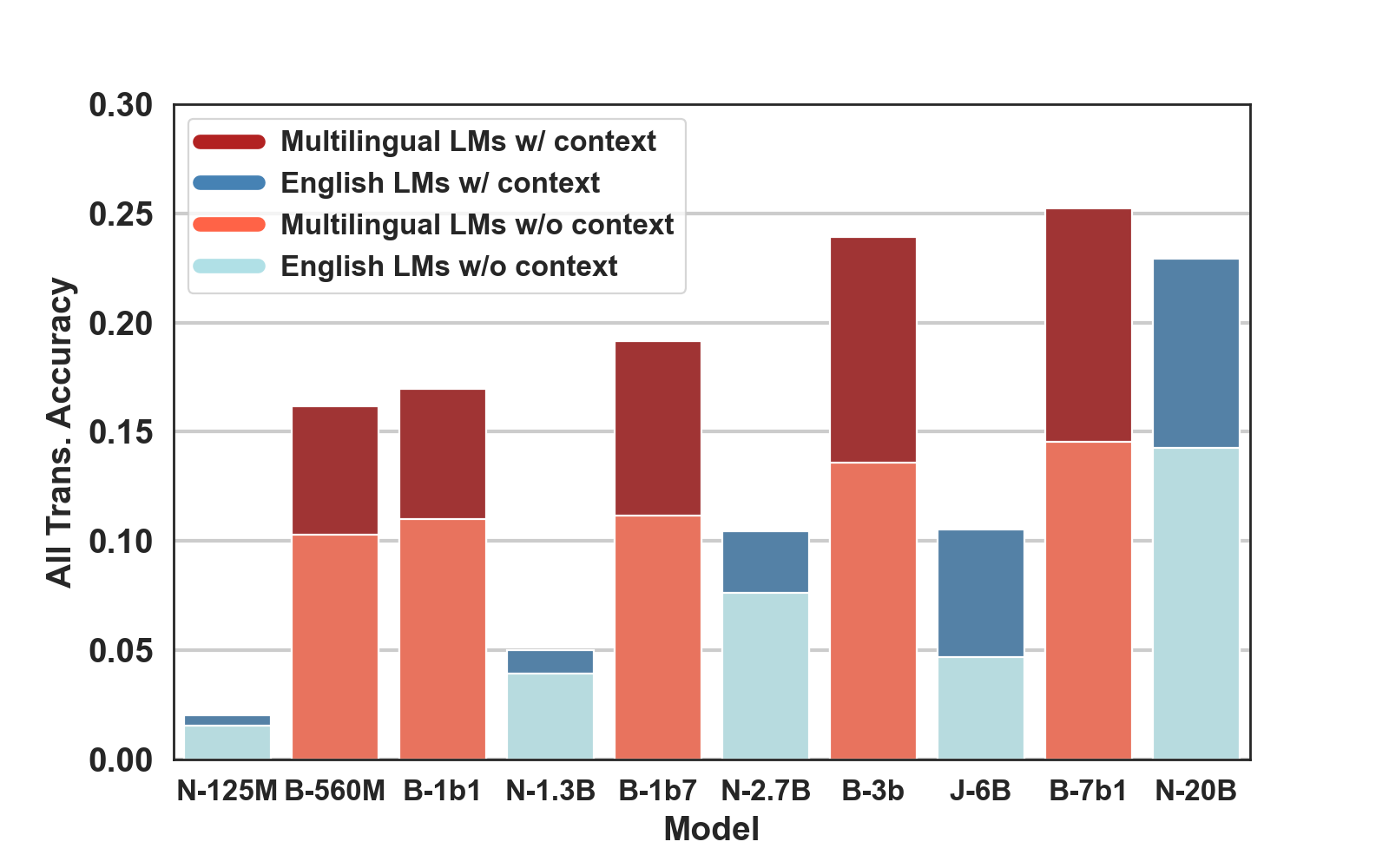} \\
        \includegraphics[width=0.48\textwidth]{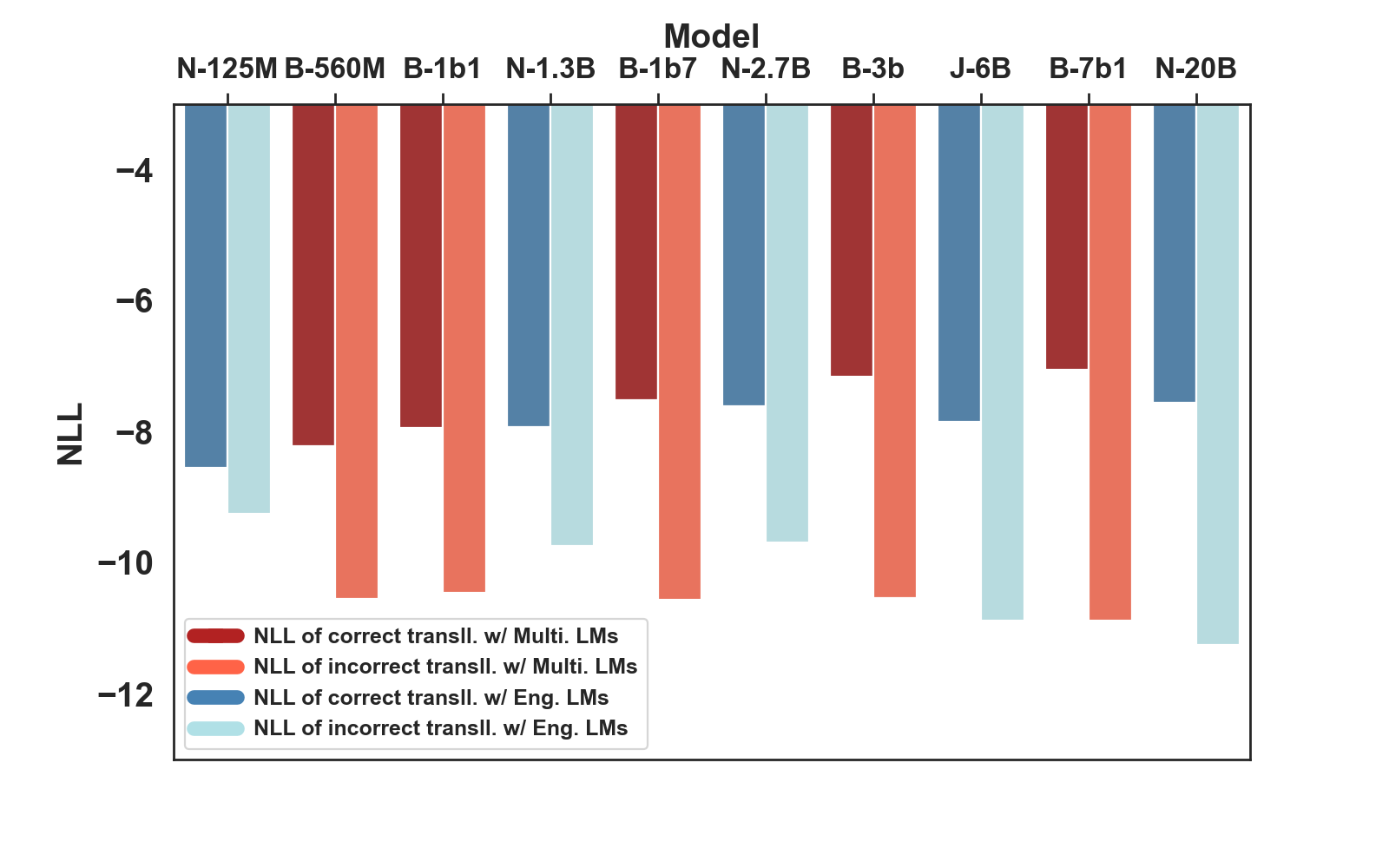} &   
        \includegraphics[width=0.48\textwidth]{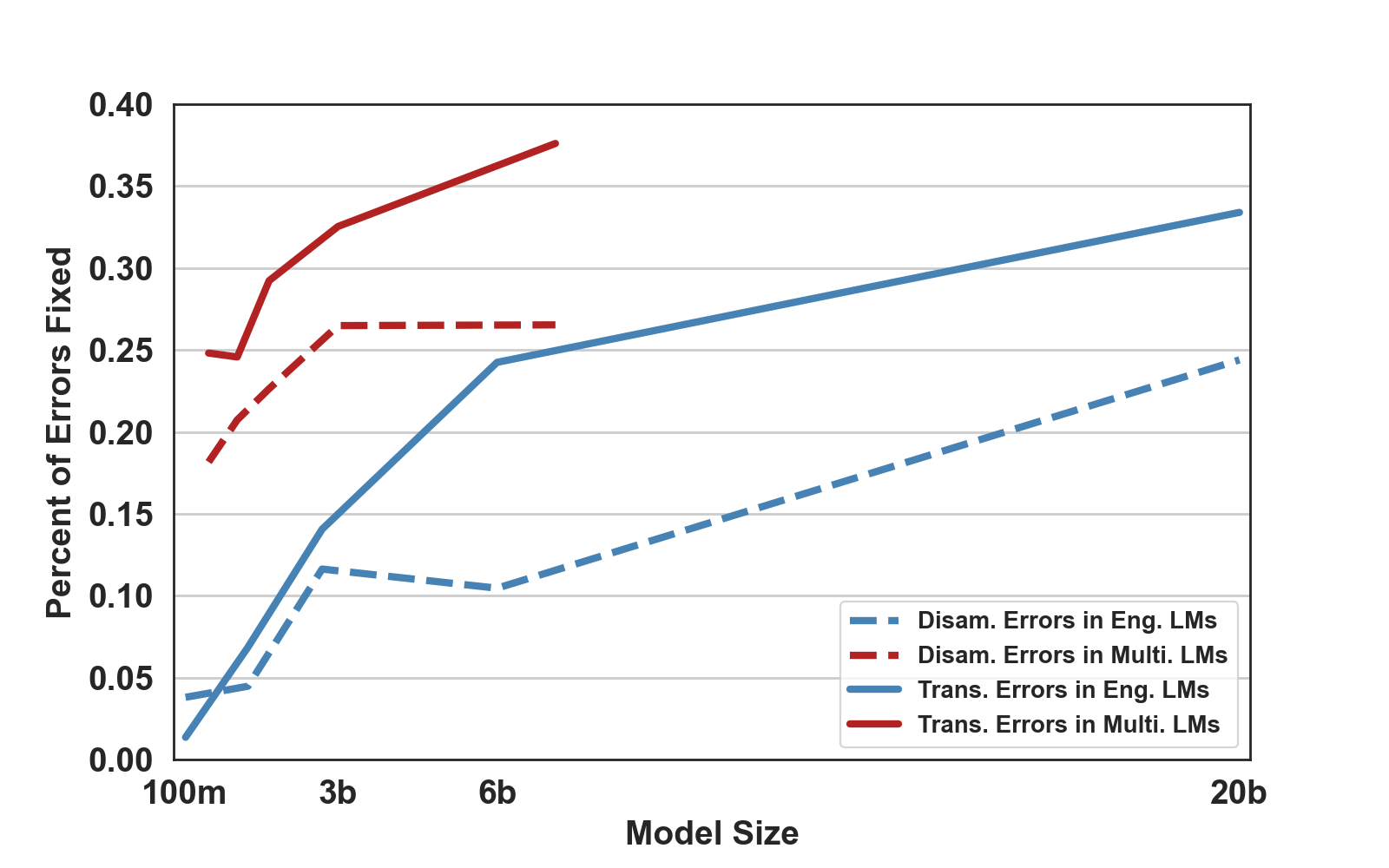} \\
    \end{tabular}
    \caption{C-WLT results for French}
    \label{fig:cwlt-french}
    \end{figure*}

\begin{figure*}
        \begin{tabular}{cc}
        \includegraphics[width=0.48\textwidth]{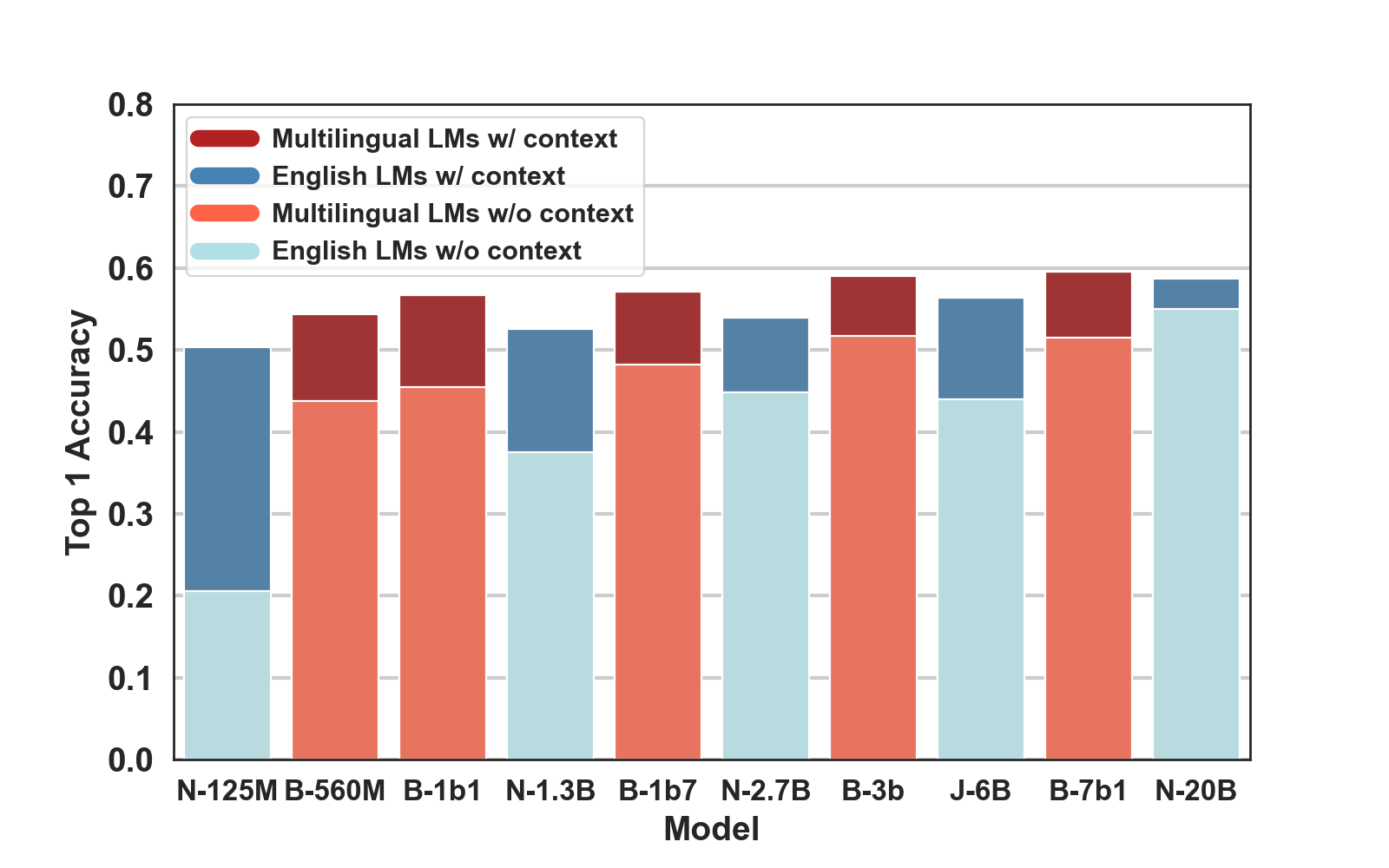} &   
        \includegraphics[width=0.48\textwidth]{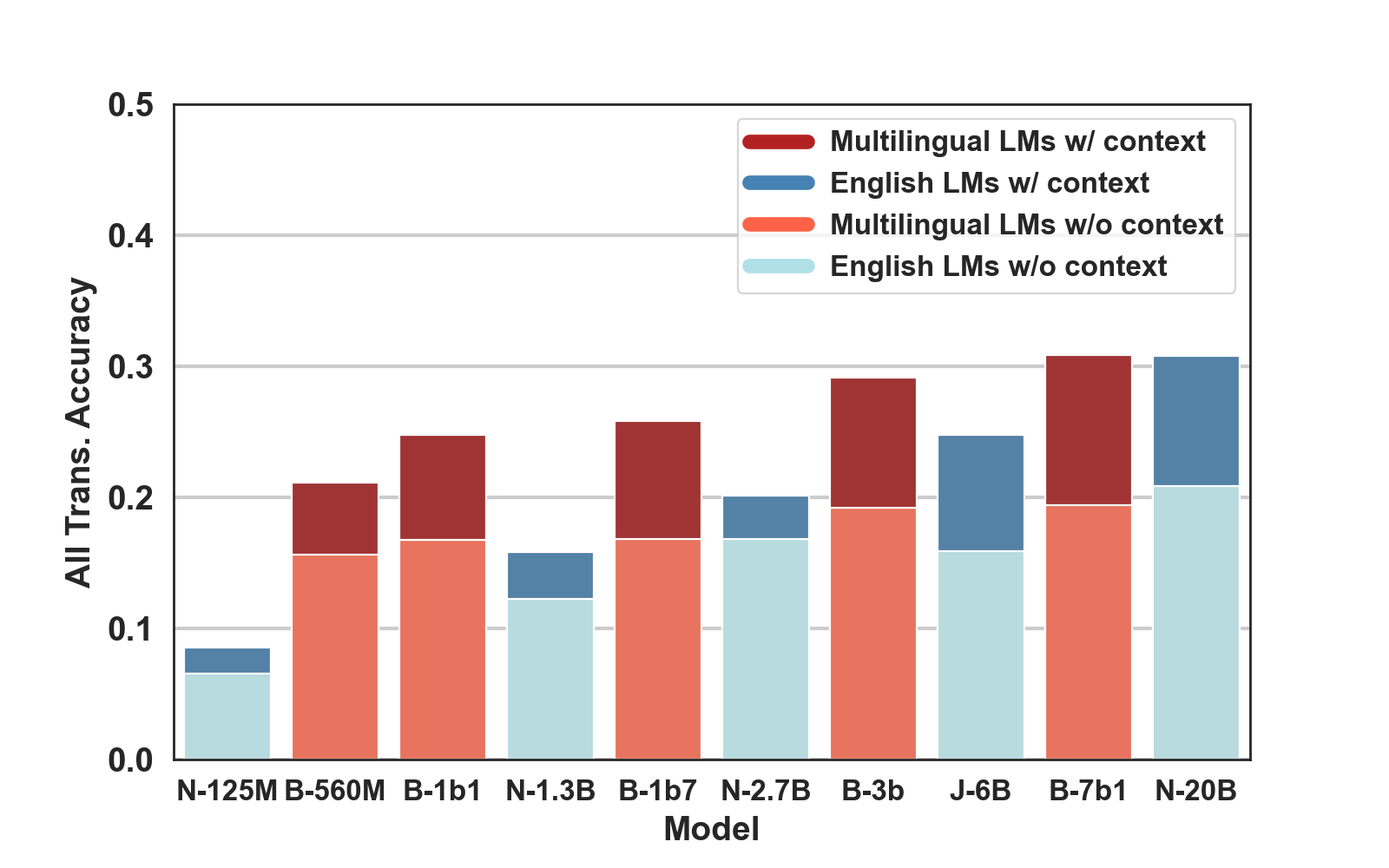} \\
        \includegraphics[width=0.48\textwidth]{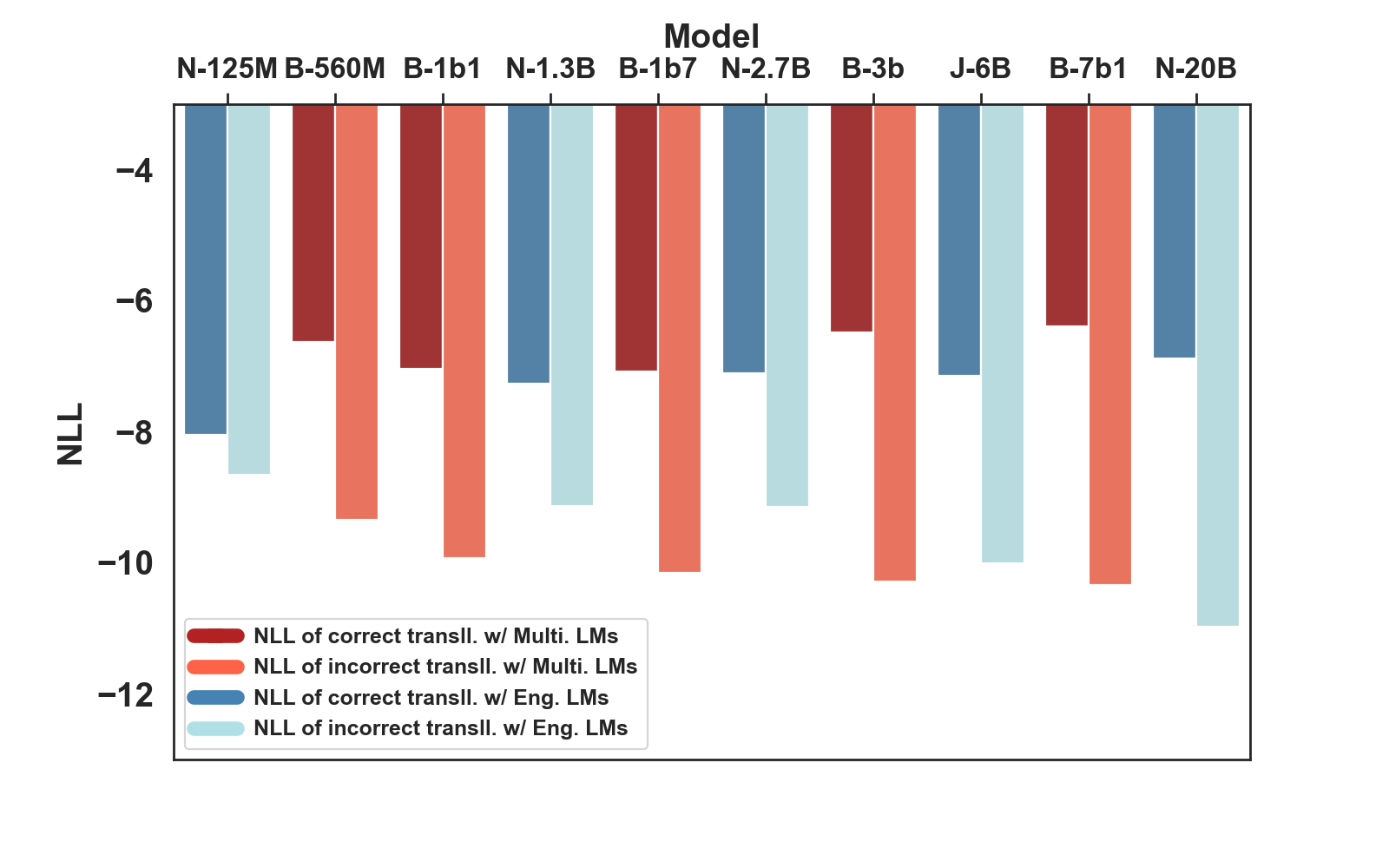} &   
        \includegraphics[width=0.48\textwidth]{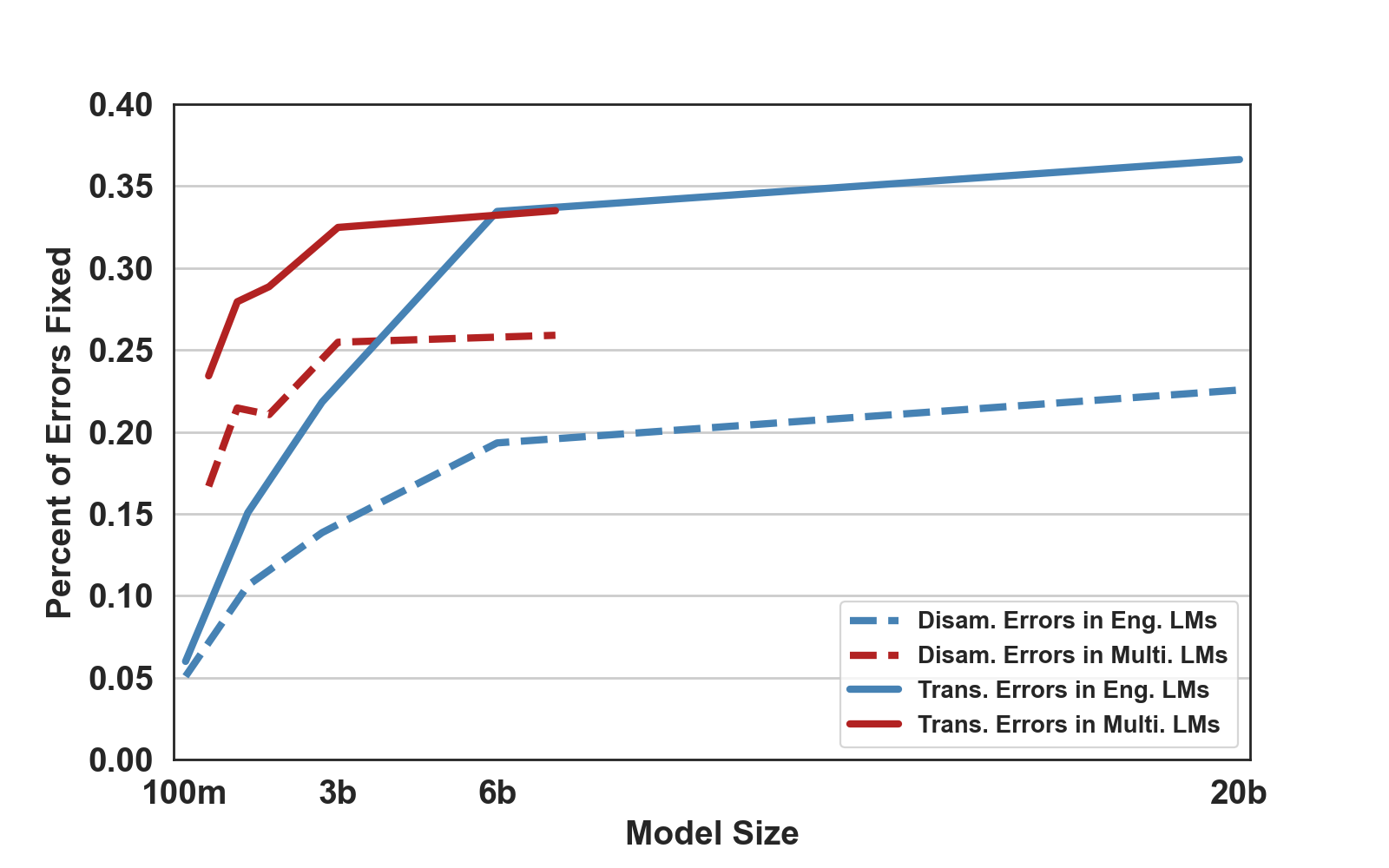} \\
    \end{tabular}
    \caption{C-WLT results for Spanish}
    \label{fig:cwlt-spanish}
    \end{figure*}

We provide the full set of results for the following experiments:
\begin{itemize}
    \item Figure \ref{fig:cwlt-chinese}: Results for the EN-ZH setting of the C-WLT experiments from Section \ref{sec:cwlt-results}.
    \item Figure \ref{fig:cwlt-french}: Results for the EN-FR setting of the C-WLT experiments from Section \ref{sec:cwlt-results}.
    \item Figure \ref{fig:cwlt-spanish}: Results for the EN-ES setting of the C-WLT experiments from Section \ref{sec:cwlt-results}.
    \item Table \ref{tab:diff_targets_bloom}: Bloom-3B and Bloom-7.1B results for the target language ablation and ensembling experiments from Section \ref{sec:ablations}.
    \item Table \ref{tab:mcs-lcs-results}: The per-language results of the MCS and LCS performance ablation from Section \ref{sec:error-analysis}.
\end{itemize}

\hfill

\hfill

\begin{table*}[]
    \centering
    \small{
    \begin{tabular}{c|c c| c c | c c}
    
         \multirow{2}{*}{\textbf{Target Lang.}} &  \multicolumn{2}{c|}{\textbf{Recall}} & \multicolumn{2}{c|}{\textbf{Jaccard Index}} & \multicolumn{2}{c}{\textbf{Delta}} \\
         & \textbf{B-3B} & \textbf{B-7.1B} & \textbf{B-3B} & \textbf{B-7.1B} & \textbf{B-3B} & \textbf{B-7.1B} \\
            \hline
        English  & 63.60 & 63.62 & 51.83 & 52.32 & 10.1 & 9.7\\
        Spanish & 69.58 & 69.86 & 52.28 & 52.31 & 15.7 & 15.6\\
        Chinese & 68.77 & 69.96 & 57.43 & 58.27 & 4.1 & 4.1\\
        Russian & 65.06 & 65.68 & 53.75 & 54.39 & 9.4 & 9.4\\
        Finnish & 55.01 & 56.52 & 47.73 & 48.73 & 6.9 & 6.5\\
        \hline
        Best Setting$^{*}$ & 68.62 & 69.45 & 57.42 & 58.24 & 8.7 & 8.2 \\
        All 5 Joint & 63.95 & 65.03 & 55.42 & 56.35 & 6.5 & 6.4\\
    \end{tabular}
    }
    \caption{The average zero-shot recalls and Jaccard Index (\%) of all 18 source languages in the XL-WSD dataset for the different target language settings for the BLOOM family PLMs. $^{*}$The best setting is the joint English, Chinese, and Russian.}
    \label{tab:diff_targets_bloom}
\end{table*}

\begin{table*}[]
    \centering
    \small
    \begin{tabular}{c|c c c c || c c c c}
    \hline
        \multirow{3}{*}{\textbf{Language}} & \multicolumn{4}{c||}{\textbf{Recall}} & \multicolumn{4}{c}{\textbf{Jaccard Index}} \\
        & \multicolumn{2}{c}{\textbf{Bloom-7.1B}} & \multicolumn{2}{c||}{\textbf{GPT-NeoX}} &\multicolumn{2}{c}{\textbf{Bloom-7.1B}} & \multicolumn{2}{c}{\textbf{GPT-NeoX}} \\
     & \textbf{MCS} & \textbf{LCS} & \textbf{MCS} & \textbf{LCS} & \textbf{MCS} & \textbf{LCS} & \textbf{MCS} & \textbf{LCS} \\
    \hline
    Basque & 79.22 & 42.25 & 71.84 & 36.24 & 72.84 & 28.48 & 65.70 & 23.41 \\
    Bulgarian & 83.54 & 56.38 & 86.79 & 60.13 & 79.04 & 42.98 & 81.18 & 45.96 \\
    Catalan & 71.89 & 50.11 & 73.13 & 48.66 & 60.37 & 32.93 & 60.45 & 30.91 \\
    Chinese & 75.82 & 49.74 & 76.81 & 52.41 & 66.56 & 35.34 & 66.45 & 36.32 \\
    Croatian & 87.63 & 50.62 & 89.01 & 54.35 & 84.32 & 38.50 & 85.26 & 41.19 \\
    Danish & 87.89 & 58.23 & 90.01 & 66.53 & 83.93 & 45.48 & 85.39 & 51.80 \\
    English & 91.20 & 61.52 & 90.51 & 61.01 & 84.63 & 46.32 & 83.90 & 45.43 \\
    Estonian & 79.33 & 44.10 & 82.53 & 51.03 & 75.03 & 33.42 & 77.73 & 36.70 \\
    French & 93.07 & 59.81 & 88.35 & 61.14 & 86.25 & 45.92 & 81.84 & 43.90 \\
    Galician & 85.83 & 66.13 & 86.54 & 64.39 & 78.37 & 47.00 & 78.56 & 46.21 \\
    German & 89.08 & 60.87 & 87.97 & 63.48 & 84.92 & 43.71 & 84.36 & 47.04 \\
    Hungarian & 81.31 & 42.24 & 84.73 & 49.50 & 77.06 & 31.30 & 79.72 & 36.84 \\
    Italian & 86.62 & 65.78 & 85.68 & 66.54 & 73.84 & 45.81 & 73.85 & 46.28 \\
    Japanese & 82.83 & 54.94 & 84.53 & 58.42 & 76.21 & 38.10 & 77.29 & 39.48 \\
    Korean & 81.98 & 42.93 & 81.47 & 41.96 & 79.31 & 32.55 & 78.71 & 32.16 \\
    Slovenian & 69.02 & 40.07 & 77.90 & 43.85 & 59.50 & 28.68 & 68.69 & 29.73 \\
    Spanish & 87.81 & 71.12 & 87.12 & 67.94 & 71.83 & 49.75 & 71.72 & 45.25 \\
    \hline
    Avg. & 83.16 & 53.93 & 83.82 & 55.74 & 75.11 & 39.19 & 76.52 & 39.92 \\
    \hline
    \end{tabular}
    \caption{Recall and Jaccard index performance of the best-ensembled WSD via C-WLT setting for the most common senses (MCS) and less common senses (LCS) of words in each evaluation language.}
    \label{tab:mcs-lcs-results}
\end{table*}

\end{document}